# A comparative evaluation and analysis of three generations of Distributional Semantic Models

**Alessandro Lenci**[1] · **Magnus Sahlgren**[2] ·
**Patrick Jeuniaux**[3] · **Amaru Cuba Gyllensten**[4] ·
**Martina Miliani**[5]

**Abstract** Distributional semantics has deeply changed in the last decades. First, predict models stole the thunder from traditional count ones, and more recently both of them were replaced in many NLP applications by contextualized vectors produced by neural language models. Although an extensive body of research has been devoted to Distributional Semantic Model (DSM) evaluation, we still lack a thorough comparison with respect to tested models, semantic tasks, and benchmark datasets. Moreover, previous work has mostly focused on task-driven evaluation, instead of exploring the differences between the way models represent the lexical semantic space. In this paper, we perform a large-scale evaluation of type distributional vectors, either produced by static DSMs or obtained by averaging the contextualized vectors generated by BERT. First of all, we investigate the performance of embeddings in several semantic tasks, carrying out an in-depth statistical

✉ Alessandro Lenci
  alessandro.lenci@unipi.it

  Magnus Sahlgren
  magnus.sahlgren@ai.se

  Patrick Jeuniaux
  patrick.jeuniaux@just.fgov.be

  Amaru Cuba Gyllensten
  amaru.cuba.gyllensten@ri.se

  Martina Miliani
  m.miliani@studenti.unistrasi.it

[1]  Università di Pisa, Pisa, Italy

[2]  AI Sweden, Stockholm, Sweden

[3]  Institut National de Criminalistique et de Criminologie, Brussels, Belgium

[4]  RISE, Stockholm, Sweden

[5]  Università per Stranieri di Siena, Siena & Università di Pisa, Pisa, Italy

analysis to identify the major factors influencing the behavior of DSMs. The results show that (i) the alleged superiority of predict based models is more apparent than real, and surely not ubiquitous and (ii) static DSMs surpass BERT representations in most out-of-context semantic tasks and datasets. Furthermore, we borrow from cognitive neuroscience the methodology of Representational Similarity Analysis (RSA) to inspect the semantic spaces generated by distributional models. RSA reveals important differences related to the frequency and part-of-speech of lexical items.

## 1 Introduction

*Distributional semantics* (Lenci, 2008, 2018; Boleda, 2020) is today the leading approach to lexical meaning representation in Natural Language Processing (NLP), Artificial Intelligence (AI), and cognitive modeling. Grounded in the *Distributional Hypothesis* (Harris, 1954; Sahlgren, 2008), according to which words with similar linguistic contexts tend to have similar meanings, distributional semantics represents lexical items with real-valued *vectors* (nowadays commonly called *embeddings*) that encode their linguistic distribution in text corpora. We refer to models that learn such representations as *Distributional Semantic Models* (DSMs).

The landscape of distributional semantics has undergone deep transformations since its outset. Three main generations of models have followed one another: (i) *count DSMs* that build distributional vectors by recording co-occurrence frequencies; (ii) *predict DSMs* that learn vectors with shallow neural networks trained to predict surrounding words; (iii) *contextual DSMs* that use deep neural language models to generate inherently contextualized vectors for each word token, and therefore radically depart from previous *static DSMs* that instead learn a single vector per word type. Across its history, the changes in distributional semantics involve the way to characterize linguistic contexts, the methods to generate word vectors, the very nature of such vectors (e.g., type vs. token ones), and the model complexity itself, which has exponentially grown especially with the last generation of deep neural DSMs, now consisting of hundreds of billions parameters and requiring equally huge amounts of computational resources for their training. In this paper, we assess the effects of such development and we report on a large-scale evaluation of the most influential past and present DSMs with a twofold aim:

1. *investigating the performance of embeddings in several semantic tasks*. We performed an in-depth statistical analysis to identify the major factors influencing the behavior of static DSMs, in particular to understand whether predict models are really superior to count ones;
2. *exploring the differences of the semantic spaces produced by both static and contextual DSMs*, zooming-in on various areas of the lexicon differing for frequency and part-of-speech (POS). This study was carried out with Representational Similarity Analysis, a technique developed in cognitive neuroscience.

All our analyses focus on *type* distributional vectors, because of their importance for linguistic tasks, even if they have been superseded by token embeddings in most NLP applications. Token vectors are attractive as they are able to capture word meaning in context. On the other hand, although contexts can induce important effects of meaning modulation, human lexical competence is also context-independent. When presented with word types, human subjects are indeed able to carry out several semantic tasks on them, such as rating their semantic similarity or group them into categories (Murphy, 2002). For instance, the fact that *dog* and *cat* belong to the same semantic category is a type-level property. This supports the hypothesis that word meanings abstract and are (at least partially) invariant from the specific contexts in which their tokens are observed. Therefore, testing type embeddings allows us to investigate their ability to capture such context-independent dimensions of lexical meaning. Besides the analysis of the type embeddings natively generated by static DSMs, we also investigate to what extent type-level semantic properties can be reconstructed from contextualized token vectors (Chronis and Erk, 2020).

This paper is organized as follows: Sect. 2 reconstructs the main evolutionary phases of distributional semantics, Sect. 3 reviews current work on DSM evaluation, Sect. 4 presents a battery of experiments to test static DSMs on intrinsic and extrinsic tasks, and Sect. 5 compares them with BERT, as a representative of the last generation of contextual DSMs. Section 6 studies their semantic spaces with Representational Similarity Analysis, and Sect. 7 discusses the significance of our findings for research on distributional semantics.[1]

## 2 From count vectors to contextual embeddings: three generations of DSMs

Many types of DSMs have been designed throughout the years (see Table 1). The first generation of DSMs dates back to the 1990s and is characterized by so-called *count models*, which learn the distributional vector of a target lexical item by recording and *counting* its co-occurrences with linguistic contexts. These can consist of documents (Landauer and Dumais, 1997; Griffiths et al., 2007) or lexical collocates, the latter in turn identified with either a "bag-of-word" window surrounding the target (Lund and Burgess, 1996; Kanerva et al., 2000; Sahlgren, 2008) or syntactic dependency relations extracted from a parsed corpus (Padó and Lapata, 2007; Baroni and Lenci, 2010). In *matrix models*, directly stemming from the Vector Space Model in information retrieval (Salton et al., 1975), target-context co-occurrence frequencies (or frequency-derived scores that are more suitable to reflect the importance of the contexts) are arranged into a *co-occurrence matrix* (or a more complex geometric object, like a tensor; Baroni and Lenci, 2010). In this matrix, target lexical items are represented with high-dimensional and sparse *explicit vectors* (Levy and Goldberg, 2014b), such that each dimension is labeled with a specific context in which the targets have been observed in the corpus. In order to improve the quality of the resulting semantic space by smoothing unseen data, removing noise, and exploiting redundancies and correlations between the linguistic contexts (Turney and Pantel, 2010), the co-occurrence matrix is typically mapped onto a reduced matrix of low-dimensional, dense vectors consisting of "latent" semantic dimensions implicit in the original distributional data. Dense vectors are generated from explicit ones by factorizing the co-occurrence matrix with Singular Value Decomposition (Landauer and Dumais, 1997) or with Bayesian probabilistic methods like Latent Dirichlet Allocation (Blei et al., 2003). A different kind of count DSMs are *random encoding models*: Instead of collecting global co-occurrence statistics into a matrix, they directly learn low-dimensional distributional representations by assigning to each lexical item a *random vector* that is incrementally updated depending on the co-occurring contexts (Kanerva et al., 2000).

---

[1] The DSMs and scripts are available on GitHub: https://github.com/Unipisa/DSMs-evaluation.

**Table 1** Some of the most influential DSMs to date

| Model name | References |
| --- | --- |
| Hyperspace Analogue of Language (HAL) | Lund and Burgess (1996) |
| Latent Semantic Analysis (LSA) | Landauer and Dumais (1997) |
| Random Indexing (RI) | Kanerva et al. (2000) |
| Dependency Vectors (DV) | Padó and Lapata (2007) |
| Topic Models | Griffiths et al. (2007) |
| Random Indexing with permutations | Sahlgren et al. (2008) |
| Distributional Memory (DM) | Baroni and Lenci (2010) |
| word2vec (CBOW, Skip-Gram) | Mikolov et al. (2013a, 2013b) |
| Global Vectors (GloVe) | Pennington et al. (2014) |
| FastText | Bojanowski et al. (2017) |
| Embeddings from Language Models (ELMo) | Peters et al. (2018) |
| Bidirectional Encoder Representations from Transformers (BERT) | Devlin et al. (2019) |

With the emergence of *deep learning* methods in the 2010s (Goodfellow et al., 2016; Goldberg, 2017), a new generation of so-called *predict models* has entered the scene of distributional semantics and competed with more traditional ones. Rather than counting co-occurrences, predict DSMs are *artificial neural networks* that directly generate low-dimensional, dense vectors by being trained as language models that learn to *predict* the contexts of a target lexical item. In this period of innovation, documents have largely been abandoned as linguistic contexts, and models have focused on window-based collocates and to a much lesser extent on syntactic ones (Levy and Goldberg, 2014b). Thanks to the deep learning wave, neural models – not necessarily deep in themselves – like word2vec (Mikolov et al., 2013a, 2013b) and FastText (Bojanowski et al., 2017) have quickly overshadowed count DSMs, though the debate on the alleged superiority of predict models has produced inconclusive results (Baroni et al., 2014; Levy et al., 2015). The only exception to the dominance of predict models in this period is represented by GloVe (Pennington et al., 2014). However, despite being a matrix DSM, the GloVe method to learn embeddings is closely inspired by neural language models.

The advent of predict DSMs has also brought important methodological novelties. Besides popularizing the expression "word embedding" as a kind of standard term to refer to distributional representations, deep learning has radically modified the scope and application of distributional semantics itself. The first generation of DSMs essentially encompassed computational methods to estimate the semantic similarity or relatedness among words (e.g., to build data-driven lexical resources). On the other hand, embeddings are nowadays routinely used in deep learning architectures to initialize their word representations. These *pretrained embeddings* allow neural networks to capture semantic similarities among lexical items that are beneficial to carry out downstream supervised tasks. Thus, distributional semantics has become a general approach to provide NLP and AI applications with semantic information. This change of perspective has also affected the approach to DSM evaluation. The previous generation of distributional semantics usually favoured *intrinsic methods* to test DSMs for their ability to model various kinds of semantic similarity and relatedness (e.g., synonymy tests, human similarity ratings, etc.). Currently, the widespread use of distributional vectors in deep learning architectures has boosted *extrinsic evaluation* methods: The vectors are fed into a downstream NLP task (e.g., part-of-speech tagging or named entity recognition) and are evaluated with the system's performance.

A further development in distributional semantics has recently come out from the research on deep neural language models. For both count and predict DSMs, a common and longstanding assumption is the building of a single, stable representation for each word type in the corpus. In the latest generation of embeddings, instead, each word token in a specific sentence context gets a unique representation. These models typically rely on a multi-layer encoder network and the word vectors are learned as a function of its internal states, such that a word in different sentence contexts determines different activation states and is represented by a distinct vector. Therefore, the distributional representations produced by these new frameworks are called *contextual embeddings* (Liu et al., 2020), as opposed to the *static* ones produced by earlier DSMs. The new generation of contextualized vectors was sparked by ELMo (Peters et al., 2018), which is based on a two-layer bidirectional LSTM. Nowadays, the most popular architectures for obtaining contextual embeddings, such as BERT (Devlin et al., 2019), consist of a stack of Transformer layers (Vaswani et al., 2017).

Generating lexical representations is not the end goal of systems like BERT or GPT (Radford et al., 2018, 2019), which are designed chiefly as general, multi-task architectures to develop NLP applications based on the technique of fine-tuning. However, since their internal layers produce embeddings that encode several aspects of meaning as a function of the distributional properties of words in texts, BERT and its relatives can also be regarded as DSMs (specifically, predict DSMs, given their language modelling training objective) that generate token distributional vectors (Mickus et al., 2020).[2] In fact, the improvements achieved by these systems in several tasks has granted huge popularity to contextual embeddings that have fast replaced static ones, especially in downstream applications. The reason for this success is ascribed to the ability of such representations to capture several linguistic features (Tenney et al., 2019) and, in particular, context-dependent aspects of word meaning (Wiedemann et al., 2019), which overcomes an important limit of static embeddings that conflate different word senses in the same type vector (Camacho-Collados and Pilehvar, 2018). In this last generation of DSMs, the contexts of the target lexical items are not selected *a priori*. Models are instead fed with huge amounts of raw texts and they learn (e.g., thanks to the attention mechanism of Transformers) which words are most related to the one that is being processed, thereby encoding in the output vectors relevant aspect of the context structure.

On the other hand, the architectures used to learn contextual embeddings have reached an unprecedented degree of complexity, even when compared to the previous generation of predict DSMs. For instance, the largest version of BERT has 340 million parameters, and some of the later models, like GPT, have hundreds of billions of them. These deep networks need to be trained on huge amounts of text and have several layers each producing a different contextualized vector encoding several types of morpho-syntactic and semantic information. Therefore, the presumed higher quality of last-generation embeddings comes to the cost of an increased amount of complexity that often obscures the understanding of their content.

## 3 Related work on DSM evaluation

There is an extensive body of research related to DSM evaluation. Broadly speaking, large scale evaluation boils down to two highly interrelated questions: (i) What is the effect of model and/or hyperparameter choice (or other factors)?; (ii) What do the evaluation metrics tell us? These questions have the following normative counterparts: (i) What is the "optimal" model and what hyperparameters should be used?; (ii) How should models be evaluated?

Most existing work is concerned with the first question and investigates the effect of model and/or hyperparameter choice. For example, Sahlgren and Lenci (2016) explore the effect of data size on model performance. Levy and Goldberg (2014a), Melamud et al. (2016), Lapesa and Evert (2017) and Li et al. (2017) study the effect of context type (i.e., window-based vs. syntactic collocates) and embedding dimension, whereas Baroni et al. (2014) and Levy et al. (2015) study the effect of modeling category (i.e., count vs. predict models). In particular, Levy et al. (2015) make the argument that model types should not be evaluated in isolation, but that using comparable hyperparameters (or tuning) is necessary for fair and informative comparison.

---

[2] Westera and Boleda (2019) instead argue that the domain of distributional semantics is limited to context-invariant semantic representations. However, context-sensitive token vectors are not an absolute novelty in the field (Erk and Padó, 2010), though they have remained a kind of sideshow until the boom of deep neural language models.

Other works instead focus on the evaluation type. Chiu et al. (2016) observe that intrinsic evaluation fails to predict extrinsic (sequence labeling) performance, stressing the importance of extrinsic evaluation. Rogers et al. (2018) introduce an extended evaluation framework, including diagnostic tests. They corroborate the findings of Chiu et al. (2016), but show that performance for other extrinsic tasks is predicted by intrinsic performance, and that there exists diagnostic tasks that predict sequence labeling performance.

However, much of this work is limited to one type or family of DSMs. For example, Levy and Goldberg (2014a), Chiu et al. (2016), Ghannay et al. (2016), Melamud et al. (2016), and Rogers et al. (2017) are all concerned solely with predict word embeddings. Whereas Bullinaria and Levy (2007, 2012) and Lapesa and Evert (2014, 2017) limit their exploration to count DSMs. Conversely, work that compares across model type boundaries has been restricted in the scope of evaluation, being singularly concerned with intrinsic evaluation (Baroni et al., 2014; Levy et al., 2015; Sahlgren and Lenci, 2016). As far as we are aware, the only large-scale comparison across model type boundaries, involving both intrinsic and extrinsic tasks, is Schnabel et al. (2015). However, they used default hyperparameters for all models, and set the embedding dimension to 50.

Recently, attention has shifted towards comparing the type embeddings produced by static DSMs with those obtained by pooling the token contextualized embeddings generated by deep neural language models (Ethayarajh, 2019; Bommasani et al., 2020; Chronis and Erk, 2020; Vulić et al., 2020). However, these works have so far addressed a limited number of tasks or a very small number of static DSMs.

This analysis of the state of the art in DSM evaluation reveals that we still lack a comprehensive picture, with respect to tested models, semantic tasks, and benchmark datasets. Moreover, the typical approach consists in measuring *how good* a given model is and which model is *the best* in a particular task. Much less attention has been devoted to exploring the differences between the way models represent the lexical semantic space. The following sections aim at filling these gaps.

## 4 Quantitative evaluation of static DSMs

A traditional DSM is defined by a set $T$ of *target words* and a set $C$ of *linguistic contexts*. The model assigns to any $t \in T$ a $n$-dimensional *distributional vector* encoding its co-occurrences with the contexts $C$. We reserve the term *word embedding* to refer only to low-dimensional, dense distributional vectors.

In this section, we present a large-scale quantitative evaluation of static DSMs, in which several models (Sect. 4.1) are tested on a wide range of semantic tasks (Sect. 4.2) with the aim of identifying the key factors affecting their performance.

**Table 2** Static DSMs and parameter settings

| Model | Context | Vector type | Dimensions |
|---|---|---|---|
| *Matrix count models* | | | |
| PPMI | window.{2,10}; syntax.{typed,filtered} | Explicit | 10,000 |
| SVD | window.{2,10}; syntax.{typed,filtered} | Embedding | 300; 2000 |
| LSA | document | Embedding | 300; 2000 |
| LDA | document | Embedding | 300; 2000 |
| GloVe | window.{2,10} | Embedding | 300; 2000 |
| *Random encoding count models* | | | |
| RI | window.{2,10} | Embedding | 300; 2000 |
| RI-perm | window.{2,10} | Embedding | 300; 2000 |
| *Predict models* | | | |
| SGNS | window.{2,10}; syntax.{typed,filtered} | Embedding | 300; 2000 |
| CBOW | window.{2,10} | Embedding | 300; 2000 |
| FastText | window.{2,10} | Embedding | 300; 2000 |

## 4.1 DSM selection

We selected 44 static DSMs defined by the combinations of three main factors: the *method* used to learn the distributional vectors (the 'model'), the *context type*, and the number of *vector dimensions* (see Table 2).

(A) *model—The type of method used to learn the vectors.* The models are representative of the major algorithms used to construct distributional representations:

    (i) *matrix count models*

        *PPMI*—this model consists in a simple co-occurrence matrix with collocate contexts (cf. below), weighted with Positive Pointwise Mutual Information (PPMI), computed as follows:

$$\mathrm{PPMI}_{\langle t,c \rangle} = \begin{cases} \mathrm{PMI}_{\langle t,c \rangle} = \log_2 \dfrac{p(t,c)}{p(t)p(c)} & \text{if } \mathrm{PMI}_{\langle t,c \rangle} > 0 \\ 0 & \text{otherwise} \end{cases} \quad (1)$$

where $p(t, c)$ is the co-occurrence probability of the target word $t$ with the collocate context $c$, and $p(t)$ and $p(c)$ are the individual target and context probabilities. As no dimensionality reduction is applied, PPMI produces high-dimensional, sparse, explicit distributional vectors;

*SVD*—like PPMI, but with low-dimensional embeddings generated with Singular Value Decomposition (SVD);

*LSA*—an implementation of Latent Semantic Analysis by Landauer and Dumais (1997) with document contexts, log entropy

weighting, and SVD. LSA was trained with Gensim (Řehůřek and Sojka, 2010);

*LDA*—a Topic Model (Griffiths et al., 2007) based on Latent Dirichlet Allocation (LDA) (Blei et al., 2003). Given a word-by-document co-occurrence matrix and $z_1, \ldots, z_k$ topics, LDA learns the word probability distribution for each topic, $\phi$, and the topic probability distribution for each document, $\theta$. Each target $t$ is represented with a topic vector $\mathbf{t} = (\phi_1, \ldots, \phi_k)$, such that $\phi_i = p(t|z_i)$. LDA was trained with Gensim;

*GloVe*—word embeddings are generated from the the co-occurrence matrix with a weighted regression model, whose learning objective is to find the vectors that minimize the squared error of the ratios of co-occurrence probabilities (Pennington et al., 2014).

(ii)   *random encoding count models*

*RI*—Random Indexing (Kanerva et al., 2000; Sahlgren, 2006) assigns random index vectors to the words in the model vocabulary, and adds the random vectors of the neighboring lexemes to the target embedding:

$$\mathbf{t}_i \leftarrow \mathbf{t}_{i-1} + \sum_{j=-n, j \neq 0}^{n} \mathbf{c}_j \qquad (2)$$

where $\mathbf{t}_i$ is the target embedding at step $i$, $n$ is the extension of the context window, $\mathbf{c}_j$ is a sparse $k$-dimensional random index vector (with $\delta$ randomly placed $+1$s and $-1$s) that acts as a fingerprint of context term $c_j$. RI uses the dynamic context weighting scheme introduced by Sahlgren et al. (2016), which incrementally changes as new contexts are encountered;

*RI-perm* – a variation of RI extended with random permutations of the random index vectors to reflect the position of context items with respect to the target word (Sahlgren et al., 2008).

(iii)   *predict models*

*SGNS*—the Skip-Gram with Negative Sampling model by Mikolov et al. (2013b), which predicts the context words based on the target item. SGNS was trained with word2vecf (Levy and Goldberg, 2014a);

*CBOW*—the Continuous Bag-Of-Words model, which predicts a target based on the context words (Mikolov et al. 2013a). CBOW was trained with the word2vec library;

*FastText*—the extension of SGNS by Bojanowski et al. (2017). FastText learns embeddings for character $n$-grams instead of word types, which are then represented as the sum of the *n*-gram embeddings.

(B) *context—The type of linguistic contexts*. They are representative of the major kinds used in distributional semantics:

(i) *undirected window-based collocates* The contexts of a target word *t* are the collocate words that occur within a certain linear distance from *t* specified by a context window [*m*, *n*], such that *m* is the number of words to the left of *t* and *n* is the number of units to the right of *t*. The collocates are *undirected*, because they are not distinguished by their position (left or right) with respect to the target. Window-based collocates do not take into account linguistic structures, since context windows are treated as bags of independent words ignoring any sort of syntactic information. Previous research has shown that the size of the context window has important effects on the resulting semantic space (Sahlgren, 2006; Bullinaria and Levy, 2007; Baroni and Lenci, 2011; Bullinaria and Levy, 2012; Kiela and Clark, 2014). Therefore, we experimented with two types of window-based DSMs:

> *window.2 (w2)*—narrow context window of size [2, 2]. Narrow windows are claimed to produce semantic spaces in which nearest neighbors belong to the same taxonomic category (e.g., *violin* and *guitar*);
> *window.10 (w10)*—wide context window of size [10, 10]. Large windows would tend to promote neighbors linked by more associative relations (e.g., *violin* and *play*).

(ii) *syntactic collocates* (only for PPMI, SVD, and SGNS) The contexts of a target *t* are the collocate words that are linked to *t* by a direct syntactic dependency (subject, direct object, etc.). Some experiments suggest that syntactic collocates tend to generate semantic spaces whose nearest elements are taxonomically related lexemes, mainly co-hyponyms (Levy and Goldberg,, 2014a). However, the question whether syntactic information provides a real advantage over window-based representations of contexts is still open (Kiela and Clark, 2014; Lapesa and Evert, 2017).

> *syntax.filtered (synf)*—dependency-filtered collocates. Syntactic dependencies are used just to identify the collocates, without entering into the specification of the contexts themselves (Padó and Lapata, 2007). Therefore, identical lexemes linked to the target by different dependency relations are mapped onto the same context;
> *syntax.typed (synt)*—dependency-typed collocates. Syntactic dependencies are encoded in the contexts, typing the collocates (e.g., nsubj-*dog*). Typing collocates with dependency relations captures more fine-grained syntactic distinctions, but on the other hand produces a much larger number of distinct contexts (Baroni and Lenci, 2010);

(iii) *documents* (only for LSA and LDA)—the contexts of a target $t$ are the documents they occur in. The use of textual contexts derives from the Vector Space Model in information retrieval, whose target semantic dimension is that of *topicality*, or *aboutness*. Documents are thus represented with their word distribution, and, symmetrically, lexical items with their distribution in documents, which can be regarded as "episodes" (Landauer and Dumais, 1997) that become associated with the words therein encountered.

(C) *dimensions—the number of vector dimensions*. The settings are 300 and 2000 for embeddings, and 10, 000 for explicit PPMI distributional vectors.

All 44 static DSMs were trained on a concatenation of ukWaC and a 2018 dump of English Wikipedia.[3] The corpus was case-folded, and then POS-tagged and syntactically parsed with CoreNLP (Manning et al., 2014), according to the Universal Dependencies (UD) annotation scheme.[4] We removed all words whose frequency was below 100. To reduce the negative effects of high frequency words, we followed Levy et al. (2015) and we created a subsampled version of the training corpus with the method by Mikolov et al. (2013b). Every lexical item $l$ whose frequency $F(l)$ was equal to or greater than the threshold $t$ was randomly removed with this probability:

$$p(l) = 1 - \sqrt{\frac{t}{F(l)}} \qquad (3)$$

where $t = 10^{-5}$. The vocabulary $V$ of the full training corpus contains 345, 232 unlemmatized word types, corresponding to more than 3.9 billion tokens. The subsampled corpus has the same vocabulary and ca. 2.2 billion tokens (a reduction of 42%). Since word2vec, word2vecf, and FastText perform subsampling natively, predict DSMs were trained on the full corpus, while count DSMs were trained on the subsampled corpus.

The DSM targets $T$ and contexts $C$ were selected as follows:

*targets—$T = V$*, for all models. Since targets are unlemmatized lexemes, every DSM assigns a distinct distributional vector to each inflected form. The reason for

---

this choice is that several benchmark datasets (e.g., analogy ones) are not lemmatized;

*contexts* – the main difference is between collocate vs. document contexts:

*collocates*—$C = V$. For syntax-based models, co-occurrences were identified using the dependency relation linking the target and the context lexeme.[5] For dependency-typed collocates, we used both direct and inverse dependencies. For instance, given the sentence *The dog barks*, we considered both the direct (*barks*, nsubj-*dog*) and inverse (*dog*, nsubj$^{-1}$-*barks*) dependencies. To reduce the very large number of context types, we applied a selection heuristics, keeping only the typed collocates whose frequency was greater than 500. For dependency-filtered collocates, both direct (*barks*, *dog*) and inverse (*dog*, *barks*) relations were used as well, but without context selection;

*documents*—$C = D$, where $D$ includes more than 8.2 million documents, corresponding to the articles in Wikipedia and ukWaC.

Both count and predict DSMs have several hyperparameters that can deeply affect the resulting semantic space. Since their complete exploration was not the goal of the present paper, we relied on the existing literature to select the settings that have been already shown to improve the model performance. Therefore, we adopted the following optimization strategies:

– we adopted the context probability smoothing by Levy et al. (2015) to compute the PPMI weights, and we raised the context frequency to the power of $\alpha = 0.75$;
– we applied a context selection heuristics for the explicit PPMI vectors, and we kept only the top 10, 000 most frequent lexemes. In fact, previous studies have proved that further expanding the number of contexts increases the training costs, without substantially improving the quality of distributional representations (Kiela and Clark, 2014; Lapesa and Evert, 2014);
– we followed Levy et al. (2015) and we discarded the singular value matrix produced by SVD, using the row vectors of **U** as embeddings of the targets;
– we trained predict DSMs with the negative sampling algorithm, using 15 negative examples (instead of the default value of 5), as Levy et al. (2015) show that increasing their number is beneficial to the model performance.

### 4.2 Tasks and datasets

We tested the 44 static DSMs on the 25 intrinsic (Sect. 4.2.1) and 8 extrinsic datasets (Sect. 4.2.2) reported in Table 3. Distributional vector similarity was measured with the cosine.

---

[5] The UD enhanced dependencies were included alongside the usual UD dependencies.

**Table 3** Datasets used in the experiments with the performance metrics

*Intrinsic evaluation*

| Dataset | Size | Metric | Dataset | Size | Metric |
|---------|------|--------|---------|------|--------|
| *Synonymy* | | | *Categorization* | | |
| TOEFL | 80 | Accuracy | AP | 402 | Purity |
| ESL | 50 | Accuracy | BATTIG | 5,231 | Purity |
| *Similarity* | | | BATTIG-2010 | 82 | Purity |
| RG65 | 65 | Correlation | ESSLLI-2008-1a | 44 | Purity |
| RW | 2,034 | Correlation | ESSLLI-2008-2b | 40 | Purity |
| SL-999 | 999 | Correlation | ESSLLI-2008-2c | 45 | Purity |
| SV-3500 | 3,500 | Correlation | BLESS | 26,554 | Purity |
| WS-353 | 353 | Correlation | *Analogy* | | |
| WS-SIM | 203 | Correlation | SAT | 374 | Accuracy |
| *Relatedness* | | | MSR | 8,000 | Accuracy |
| WS-REL | 252 | Correlation | GOOGLE | 19,544 | Accuracy |
| MTURK | 287 | Correlation | SEMEVAL-2012 | 3,218 | Accuracy |
| MEN | 3,000 | Correlation | WORDREP | 237,409,102 | Accuracy |
| TR9856 | 9,856 | Correlation | BATS | 98,000 | Accuracy |

*Extrinsic evaluation*

| Dataset | Training size | Test size | Task | Metric |
|---------|---------------|-----------|------|--------|
| CONLL-2003 | 204,566 | 46,665 | Sequence labeling (POS tagging) | F-measure |
| CONLL-2003 | 204,566 | 46,665 | Sequence labeling (chunking) | F-measure |
| CONLL-2003 | 204,566 | 46,665 | Sequence labeling (NER) | F-measure |
| SEMEVAL-2010 | 8,000 | 2,717 | Semantic relation classification | Accuracy |
| MR | 5,330 | 2,668 | Sentence sentiment classification | Accuracy |
| IMDB | 25,000 | 25,000 | Document sentiment classification | Accuracy |
| RT | 5,000 | 2,500 | Subjectivity classification | Accuracy |
| SNLI | 550,102 | 10,000 | Natural language inference | Accuracy |

### 4.2.1 Intrinsic tasks

For the intrinsic evaluation, we used the most widely used benchmarks in distributional semantics, grouped into the following semantic tasks:

*Synonymy* The task is to select the correct synonym to a target word from a number of alternatives. A DSM makes the right decision if the correct word has the highest cosine among the alternatives. The evaluation measure is *accuracy*, computed as the ratio of correct choices returned by a model to the total number of test items.

*Similarity* The task is to replicate as closely as possible human ratings of semantic similarity, as a relation between words sharing similar semantic attributes (e.g., *dog* and *horse*), hence the name of *attributional* (or taxonomic) similarity (Medin et al., 1993). The evaluation measure is the *Spearman rank correlation* ($\rho$) between cosine similarity scores and ratings.

*Relatedness* The task is to model human ratings of semantic relatedness. The latter is a broader notion than similarity (Budanitsky and Hirst, 2006), since it depends on the existence of *some* semantic relation or association between two words (e.g., *horse* and *saddle*). The evaluation measure is Spearman correlation;

*Categorization* The task consists in grouping lexical items into semantically coherent classes. We operationalized it with *K-means clustering*, and we measured the DSM performance with *purity* (Baroni and Lenci, 2010):

$$\text{purity} = \frac{1}{n} \sum_{r=1}^{k} \max_i (n_r^i) \tag{4}$$

where $n_r^i$ is the number of items from the $i$th true (gold standard) class that are assigned to the $r$th cluster, $n$ is the total number of items and $k$ the number of clusters. In the best case (perfect clusters), purity will be 1, while, as cluster quality deteriorates, purity approaches 0.

*Analogy completion* The task consists in inferring the missing item in an incomplete analogy $a : b = c : ?$ (e.g., given the analogy *Italy : Rome = Sweden : ?*, the correct item is *Stockholm*, since it is the capital of Sweden). The analogy task targets *relational similarity* (Medin et al. 1993), since the word pairs in the two members of the analogy share the same semantic relation. We addressed analogy completion with the *offset method* popularized by Mikolov et al. (2013c), which searches for the target lexeme $t$ that maximizes this equation:

$$\text{argmax}_{t \in T^*} (\text{sim}_{\cos}(\mathbf{t}, \mathbf{c} + \mathbf{b} - \mathbf{a})) \tag{5}$$

where $T^*$ is the set of target lexemes minus $a$, $b$, and $c$. The evaluation measure is accuracy, as the percentage of correctly inferred items.[6]

The model coverage of the datasets used in all tasks is very high (mean 98%, standard deviation 3.3%). The intrinsic performance measures were obtained by running an extended version of the Word Embeddings Benchmarks (Jastrzbski et al., 2017).[7]

### 4.2.2 Extrinsic tasks

For the extrinsic evaluation, distributional vectors were fed into supervised classifiers for the following tasks (the size of the training and test sets is reported in Table 3):

*Sequence labeling (POS tagging, chunking, and NER)* The task is to correctly identify the POS, chunk, or named entity (NE) tag of a given token. The model is a multinomial logistic regression classifier on the concatenated word vectors of a context window of radius two around – and including – the target token. The performance metric is the F-measure.

---

[6] Levy and Goldberg (2014b) proposed a variant of the offset method called *3CosMult*, which they show to obtain better performances. However, since we are not interested in the best way to solve analogies but rather in comparing different DSMs on such task, we have preferred to use the original approach.

[7] https://github.com/kudkudak/word-embeddings-benchmarks

*Semantic relation classification* The task is to correctly identify the semantic relation between two target nominals. The model is a Convolutional Neural Network (CNN). The performance metric is accuracy.

*Sentence sentiment classification* The task is to classify movie reviews as either positive or negative. The binary classification is carried out with the CNN by Kim (2014). The performance metric is accuracy.

*Document sentiment classification* The task is to classify documents as either positive or negative. The classifier is a Long Short–Term Memory (LSTM) network with 100 hidden units. The performance metric is accuracy.

*Subjectivity classification* The task is to identify whether a sentence is a subjective movie review or an objective plot summary. The model is a logistic regression classifier and input sentences are represented with the sum of their constituent word vectors. The performance metric is accuracy.

*Natural language inference* Given a premise–hypothesis pair, the task is to identify whether the premise entails, contradicts, or neither entails nor contradicts the hypothesis. The model is based on two LSTMs, one for the premise and one for the hypothesis, whose final representations are concatenated and fed into a multi-layer softmax classifier. The performance metric is accuracy.

The eight extrinsic performance measurements were computed with the Linguistic Diagnostic Toolkit (Rogers et al., 2018).[8]

## 4.3 Analyses of the static DSMs

Each of the 44 static DSMs was tested on the 33 datasets, for a total of 1,452 experiments. Table 4 contains the best score and model for each benchmark. Top performances are generally close to or better than state-of-the-art results for each dataset, and replicate several trends reported in the literature. For instance, the similarity datasets RW and SL-999 are much more challenging than WS-353 and especially MEN. In turn, the verb-only SV-3500 is harder than SL-999, in which nouns represent the lion's share. Coming to the analogy completion task, MSR and GOOGLE prove to be fairly easy. As observed by Church (2017), the performance of the offset method drastically drops with the other datasets. Moreover, it does not perform evenly on all analogy types. The top score by the FastText.w2.300 model is 0.73 on the syntactic subset of GOOGLE analogies, and 0.69 on the semantic one. Differences are much stronger in BATS: The best performance on inflection and derivation analogies is 0.43, against 0.16 on semantic analogies, and just 0.06 on analogies based on lexicographic relations like hypernymy.

A crucial feature to notice in Table 4 is that there is no single "best model". In fact, the performance of DSMs forms a very complex landscape, which we explore here with statistical analyses that focus on the following objectives: (i) determining the role played by the three main factors that define the experiments – model, context, and vector dimensions – and their possible interactions; (ii) identifying

[8] http://ldtoolkit.space.

**Table 4** Best score and static DSM for each dataset

*Intrinsic evaluation*

| Dataset | Score | Model | Dataset | Score | Model |
|---------|-------|-------|---------|-------|-------|
| *Synonymy* | | | *Categorization* | | |
| TOEFL | 0.92 | FastText.w2.2000 | AP | 0.75 | SVD.synt.300 |
| ESL | 0.78 | SVD.synt.2000 | BATTIG | 0.48 | SGNS.synt.300 |
| *Similarity* | | | BATTIG-2010 | 1.00 | SVD.synf.300 |
| RG65 | 0.87 | GloVe.w10.2000 | ESSLLI-2008-1a | 0.95 | SVD.synf.300 |
| RW | 0.48 | FastText.w2.300 | ESSLLI-2008-2b | 0.92 | SGNS.w2.2000 |
| SL-999 | 0.49 | SVD.synt.2000 | ESSLLI-2008-2c | 0.75 | SGNS.w2.2000 |
| SV-3500 | 0.41 | SVD.synt.2000 | BLESS | 0.88 | SVD.synf.2000 |
| WS-353 | 0.71 | CBOW.w10.300 | *Analogy* | | |
| WS-SIM | 0.76 | SVD.w2.2000 | SAT | 0.34 | SVD.synt.300 |
| *Relatedness* | | | MSR | 0.68 | FastText.w2.300 |
| WS-REL | 0.66 | CBOW.w10.300 | GOOGLE | 0.76 | FastText.w2.300 |
| MTURK | 0.71 | FastText.w2.300 | SEMEVAL-2012 | 0.38 | SVD.synt.300 |
| MEN | 0.79 | CBOW.w10.300 | WORDREP | 0.27 | FastText.w2.300 |
| TR9856 | 0.17 | FastText.w2.300 | BATS | 0.29 | FastText.w2.300 |

*Extrinsic evaluation*

| Dataset | Task | Score | Model |
|---------|------|-------|-------|
| CONLL-2003 | Sequence labeling (POS tagging) | 0.88 | SGNS.synt.2000 |
| CONLL-2003 | Sequence labeling (chunking) | 0.89 | SGNS.synt.300 |
| CONLL-2003 | Sequence labeling (NER) | 0.96 | SGNS.w2.2000 |
| SEMEVAL-2010 | Semantic relation classification | 0.78 | SGNS.w2.2000 |
| MR | Sentence sentiment classification | 0.78 | SGNS.w2.2000 |
| IMDB | Document sentiment classification | 0.82 | FastText.w2.300 |
| RT | Subjectivity classification | 0.91 | FastText.w2.2000 |
| SNLI | Natural language inference | 0.70 | CBOW.w2.2000 |

which DSMs are significantly different from each other; (iii) check how the task type influences the performance of the DSMs.

Here, we investigate the role played by the different factors in the experiments. First, we present the results of a global analysis (Sect. 4.3.1), and then we explore the behavior of the DSMs in the different semantic tasks (Sect. 4.3.2). The global analysis poses the problem of having a response (or dependent) variable that is homogeneous across tasks in which performance is evaluated according to different metrics (see Table 3). It is evident that an accuracy value of 0.5 has a very different meaning compared to a correlation coefficient equal to 0.5. In order to address this issue, we defined as response variable the position (*rank*) of a DSM in the performance ranking of each task. If a model A has a higher accuracy than a model

B in a task and a higher correlation than B in another task, in both cases we can say that A is "better" than B. Therefore, given a task $t$, we ordered the performance scores on $t$ in a decreasing way, and each DSM was associated with a value corresponding to its rank in the ordered list (e.g., the top DSM has rank 1, the second best scoring DSM has rank 2, and so on). The response variable is therefore defined on a scale from 1 to 44 (the number of DSMs tested on each task), in which *lower values correspond to better performances*. This conversion causes a loss of information on the distance between the scores, but normalizes the metrics both in terms of meaning and in terms of the statistical characteristics of their distribution.

### 4.3.1 Global analysis

The behavior of DSMs greatly varies accross tasks, as illustrated in Fig. 7 in Appendix A, which presents the global rank distribution of the 44 DSMs in all the 33 benchmarks. Model is the primary factor in determining the score of the experiments, context has a more contained and nuanced role, while the role of dimensions is marginal. This is confirmed by the Kruskal-Wallis rank sum non-parametric test: models (H = 854.27, df=9, p $<$ .001**) and contexts (H = 229.87, df=4, p $<$ .001**) show significant differences, while the vector dimension levels do not (H = 3.14, df=2, p = .21), as also shown in the boxplots in Fig. 1c. The only cases in which vector size matters are RI models, whose 2, 000-dimensional embeddings tend to perform better than 300-dimensional ones.

Looking at Fig. 1a, we can observe that there are three major groups of models: (i) the best performing ones are the predict models SGNS, CBOW and FastText, (ii) closely followed by the matrix models GloVe, SVD and PPMI, (iii) while the worst performing models are RI, the document-based LSA, and in particular LDA. Dunn's tests (with Bonferroni correction) were carried out to identify which pairs of models are statistically different. The p-values of these tests reported in Table 7 in Appendix A draw a very elaborate picture that can not be reduced to the strict contrast between predict and count models: (i) no difference exists between SGNS and FastText; (ii) GloVe does not differ from CBOW and SVD, and the latter two are only marginally different; (iii) interestingly, the PPMI explicit vectors do not differ from their implicit counterparts reduced with SVD and only marginally differ from GloVe; (iv) LSA does not differ from RI and RI-perm, which in turn do not differ from LDA. With regard to context types, the best scores are for syntax-based ones, either filtered or typed, while document is clearly the worst (Fig. 1b). However, we note that syntax.filtered is equivalent to syntax.typed, and the latter does not differ from window.2. On the other hand, window.10 and document are significantly different from all other context types (see Table 8 Appendix A).

As a further analysis, we fit a regression tree model (Witten and Frank, 2005) to the experiment performance scores (*dependent variable*) as a function of context, model, and vector dimensions (*independent variables* or *explanatory factors*). Regression tree analysis partitions the set of all the experiments in subsequent steps, identifying for each of them the independent variable and the combinations of its modalities that best explain the variability of the performance score. The tree growth process is blocked when the information gain becomes negligible. The output tree of our statistical model is presented in Fig. 2. The variables used to divide the experiments (nodes) are highlighted, step by step, together with the modalities with respect to which the partition is made. The tree leaves are the most uniform subgroups in terms of variability of the dependent variable with respect to the explanatory factors.

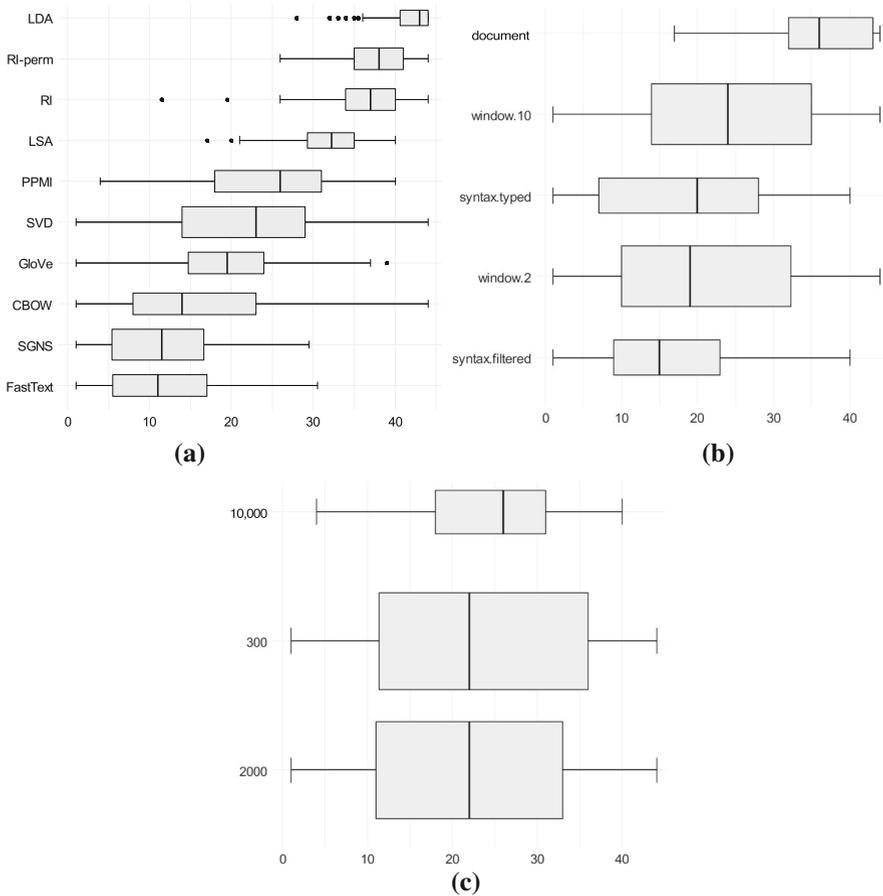

**Fig. 1** Global DSM performance: **a** per model type, **b** per context type, **c** per vector dimensions

The statistical model fit measured by the $R^2$ coefficient is 0.65, which means that model, context, and vector dimensions are able to explain just 65% of the overall variability of the DSM performance (55% of which is explained by the first two partitions, both associated with model type). The regression tree confirms the relevant role played by the model factor. In the first partition, LDA, LSA, RI and RI-perm are identified on the right branch of the tree, whose leaves are characterized by the highest average score ranks, in particular for LDA (41.3), and RI and RI-perm with size 300 (38.99). On the left branch, we find SGNS and FastText, which in the case of syntax.filtered and window.2 contexts have the lowest average score ranks (10.2). An interaction between model and context exists for CBOW, PPMI and SVD, which have worse performances (i.e., higher score ranks) with window.10 contexts.

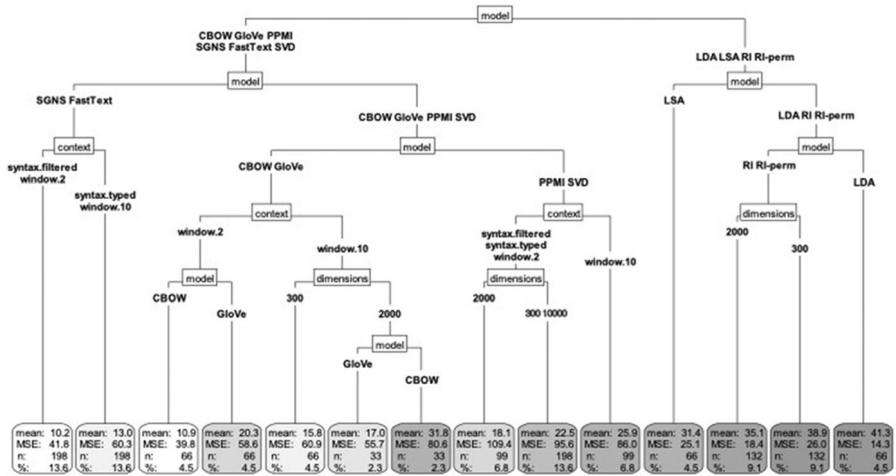

**Fig. 2** Regression tree of DSMs

### 4.3.2 Analysis by type of semantic task

DSM performance greatly varies depending on the benchmark dataset and semantic task. This is already evident from the spread out data distribution in the boxplots in Fig. 7, and is further confirmed by the regression tree analysis: After introducing task type and evaluation type (intrinsic vs. extrinsic) as further predictors, the explained data variability increases from 65% to 72%. Therefore, the behavior of DSMs is strongly affected by the semantic task they are tested on. In this section, we investigate this issue by analysing the performance of the different model and context types in the six tasks in which the datasets are grouped: synonymy, similarity, relatedness, categorization, analogy, and extrinsic tasks (see Table 3).

Figure 3 reports the per-task rank distribution of model types. In general, the best performances are obtained by SGNS and FastText, and the worst ones by LDA, RI, Ri-perm, and LSA. Instead, PPMI, SVD, GloVe and CBOW produce more intermediate and variable results: They are equivalent to, or better than, the top models in some cases, worse in others. Table 9 in Appendix A shows the model pairs whose performance is statistically different for each task, according to Dunn's test.

It is important to notice that in several cases the differences between models are actually non significant: (i) CBOW never differs from SGNS and FastText; (ii) SVD differs from predict models only in the analogy and extrinsic tasks (but for FastText in the relatedness task too), and differs from PPMI for similarity and extrinsic tasks;

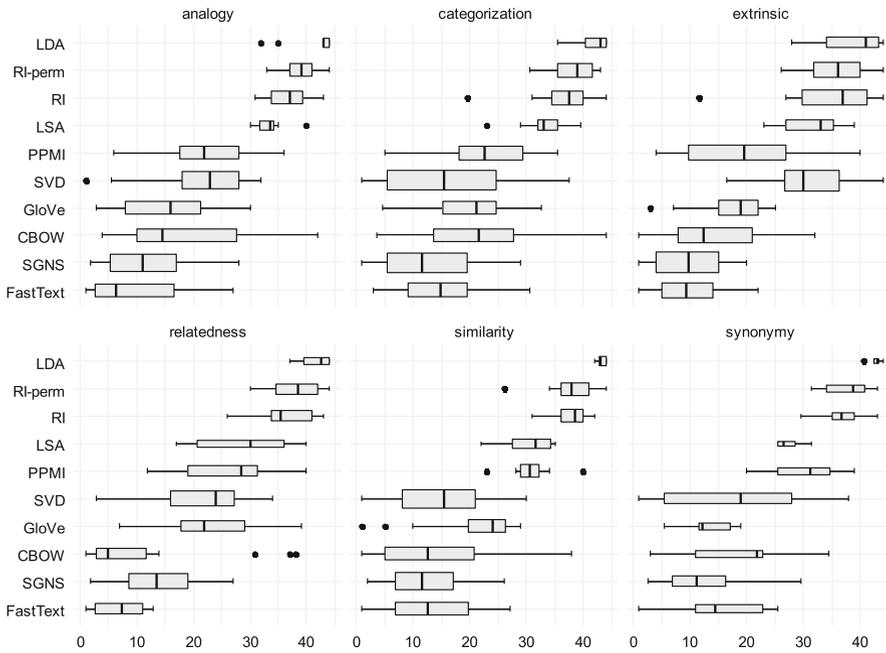

**Fig. 3** DSM performance per semantic task: rank by model type

(iii) GloVe never differs from SGNS and CBOW, and differs from FastText only for relatedness. Interestingly, GloVe differs neither from PPMI explicit vectors, nor from SVD (apart from the extrinsic task). If we exclude LDA, LSA and RI, which are systematically underperforming, models mostly differ in the extrinsic task (40% of the overall cases), with the predict ones having a clear edge over the others. Conversely, the performances in synonymy and categorization tasks are more similar across models, with only the 6% of significant pairwise differences.

Figure 4 shows the per-task distribution of the various context types. Apart from document contexts, which have the worst performance in every task, the other contexts almost never produce significant differences, as confirmed by the results of the Dunn's tests reported in Table 10 in Appendix A. The only exception is the categorization task, in which syntax-based collocates achieve significantly better performances than window-based ones. Moreover, syntax.filtered improves over window.10 in the similarity and the analogy tasks.

A further aspect we investigated is the correlation of DSM performance across datasets. Results are illustrated in Fig. 8 in Appendix A. The dot size is proportional to the Spearman correlation between the 33 datasets with respect to the performance of the 44 DSMs: The higher the correlation between two datasets the more DSMs tend to perform similarly on them. Intrinsic evaluation benchmarks are strongly correlated with each other. The only exceptions are TR9856 and ESSLLI-2008-2b. The former case probably depends on some idiosyncrasies of the dataset, as suggested by the fact that the top performance on TR9856 is significantly lower than the ones on other relatedness benchmarks (see Table 4). ESSLI-2008-2b instead focuses on concrete-abstract categorization, a task that targets a unique semantic dimension among the tested benchmarks. Intrinsic datasets are strongly correlated with extrinsic ones too, except for the sequence labeling tasks (POS tagging, chunking and NER), which have weaker correlations with the other extrinsic datasets as well.

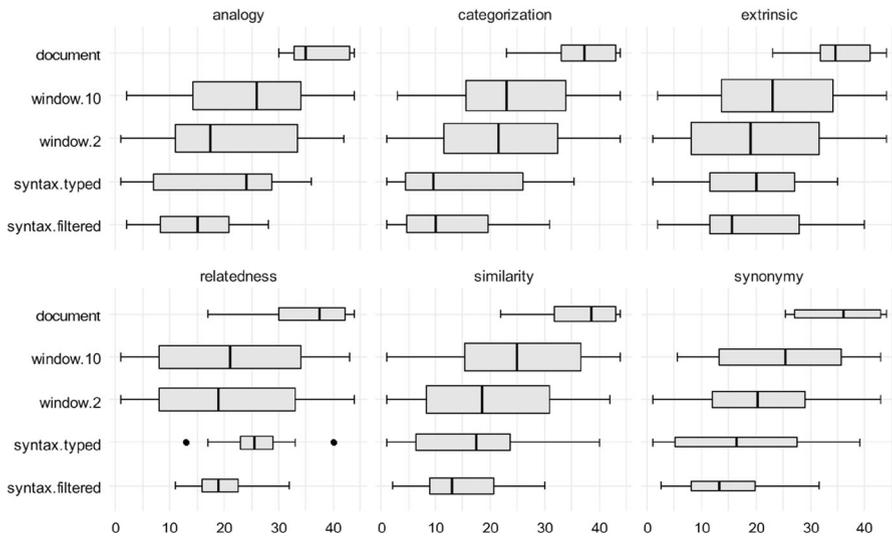

**Fig. 4** DSM performance per semantic task: rank by context type

### 4.3.3 Interim summary

The statistical analyses of static DSMs show that the model type, more than the choice of linguistic context, affects their performance, which is also strongly influenced by the type of semantic task. Overall, the most effective distributional representations are the embeddings produced by predict models and SVD or GloVe count DSMs, using either window-based or syntactic collocates as contexts. Interestingly, explicit high-dimensional PPMI vectors are also very competitive, especially in intrinsic tasks. This fact is particularly important, because of the higher interpretability of the dimensions of such distributional vectors. Among the top models, the variability across tasks is very strong, and in most cases differences do not reach statistical significance. This also confirms that the alleged superiority of predict DSMs over count ones is far from being as clear-cut as the scientific literature would suggest.

## 5 Static DSMs vs. BERT

In this section, we compare the performance of static DSMs and contextual embeddings. New neural language models are continuously appearing (Han et al., 2021): They differ for the network architecture, training task and data, number of parameters, etc. A thorough analysis of the contextual embeddings generated by these models is beyond the limits of the present paper. Therefore, we decided to focus our study on BERT (Devlin et al., 2019), which is the most famous and widely used representative of the new generation of contextual embeddings. Moreover, previous studies have shown that BERT is generally able to outperform more complex models, like GPT-2 (Ethayarajh, 2019; Bommasani et al., 2020; Vulić et al., 2020).

Different methods to derive type embeddings from contextualized token vectors can be used (Chronis and Erk, 2020). Following Bommasani et al. (2020), we computed the type embedding of a target word $t$ by averaging its BERT representations:

$$\mathbf{t} = mean(\mathbf{t}_{c_1}, \ldots, \mathbf{t}_{c_n})\tag{6}$$

where each context $c_i$ corresponds to a sentence $s \in S$ where the target occurs, and $\mathbf{t}_{c_i}$ is the token embedding for $t$ in the context $c_i$. $S$ is a random sample of sentences extracted from the same corpus used to train the static DSMs, to enhance model comparability. The minimum sentence length was set at 4 tokens, and the maximum length at 21 tokens. We extracted a maximum of 10 sentences for each target (Vulić et al., 2020), but in a few cases less sentences were found in the corpus. The selected sentences were then fed into the pretrained `bert-large-uncased` model.[9]

The large version of BERT has 24 layers, each generating embeddings of 1024 dimensions. Since BERT layers encode different linguistic features (Jawahar et al., 2019), we defined three kinds of type BERT embeddings:

*BERT.F4*—the sum of the embeddings from the first four layers;[10]
*BERT.L4*—the sum of the embeddings from the last four layers;
*BERT.L*—the embeddings from the last layer.

As they were trained on different corpora, 43.3% of the target lexemes of the static DSMs are not part of the BERT vocabulary. BERT treats out-of-vocabulary tokens in two different ways: By splitting them into subwords, or by considering the token as "unknown" (i.e., it is labeled as `[UNK]`). In the first case, like Bommasani et al. (2020) and Vulić et al. (2020), we summed the embeddings that encode the target subwords. In the second case, the targets labelled as `[UNK]` are treated like any other token. In fact, the context where a token occurs heavily affects its representation in BERT, and "unknown" words make no exception. Therefore, the resulting embedding of an `[UNK]` token is eventually quite similar to the embedding of a word in the BERT vocabulary that occurs in the same context.

BERT type embeddings were evaluated in the intrinsic setting only (cf. Sect. 4.2.1). In fact, the extrinsic tasks involve in-context lexemes, and token embeddings have already been shown to have the edge over type ones in such case. Our goal is instead to evaluate how BERT type embeddings represent the semantic properties of out-of-context words in the datasets used in the intrinsic evaluation semantic tasks.

---

[9] The tokenization and the embedding extraction were performed without stripping the accents from each target word, using the Hugging Face Python library (Wolf et al., 2020).

[10] Actually from layer 2 to layer 5, as we skipped the first layer, which corresponds to the context-independent embedding of the input token.

Previous research reported that type embeddings derived from contextual DSMs, in particular BERT, generally outperform the ones produced by static DSMs (Ethayarajh, 2019; Bommasani et al., 2020; Vulić et al., 2020). However, the picture emerging from our experiments is rather the opposite. As illustrated in Table 5, static DSMs have a clear edge over BERT in most out-of-context semantic tasks and datasets, with the only exception of SL-999, MSR, WORD-REP, and BATS, but differences in the latter datasets are not significant. In some cases, one of the BERT models is at most able to get close to the top-scoring static DSM, but in all other datasets BERT lags behind up to 20 points. The good performance on SL-999 suggests that BERT type embeddings are able to capture semantic similarity fairly well, and in fact this is the task in which BERT performances are closest to the ones of static DSMs.

In the analogy task, the highest score is obtained by BERT on MSR, but this dataset only contains syntactic analogies. In GOOGLE, BERT.L4 achieves 0.66 accuracy (0.76 for static DSMs), but its performance is indeed much better in the syntactic subset (0.71 BERT.L4; 0.73 FastText.w2.300), than in the semantic one (0.55 BERT.L4; 0.69 FastText.w2.300). The situation with BATS is exactly the same, with BERT performance on morphosyntactic analogies (0.52 BERT.L4; 0.43 FastText.w2.300) being much higher than on the semantic subset (0.11 BERT.L4; 0.11 FastText.w2.300). The generally better performance of static embeddings in the analogy task is consistent with the results reported by Ushio et al. (2021).

One key aspect of deep learning models like BERT is understanding what information is encoded in their various layers. In Table 5, we can notice the gain brought by averaging the embeddings from several layers, as already shown by Vulić et al. (2020). BERT.L, which is derived from the last and most contextualized layer, is globally the worst performing model. This is in line with the analysis by Carlsson et al. (2021), who demonstrate that masked language modeling introduces a bias onto the final layers of the models, which become less suitable to use for general semantic tasks. Carlsson et al. (2021) demonstrate that this bias can be corrected using the Contrastive Tension technique. Bommasani et al. (2020) always find best performances at earlier layers, whereas our picture is more varied: BERT.L4 generally performs better than BERT.F4 in synonymy, similarity and categorization tasks, but with relatedness datasets the behavior is reversed. This suggests that the last layers might encode semantic dimensions useful to capture attributional similarity. BERT behavior in the analogy completions task is instead more diversified, but the highest score in MSR is obtained by BERT.F4, confirming that the first layers tend to encode morphosyntactic information (Jawahar et al. 2019; Tenney et al. 2019).

The outcome of these experiments strongly contrasts with those reported in the literature. What is the reason of such difference? The performances of our BERT models are essentially in line with or better than the ones obtained by previous works. For instance, the best BERT model in Bommasani et al. (2020) (see Table 1 in their paper) achieves correlations of 0.52 in SL-999 and 0.43 in SV-3500, against

**Table 5** Best static DSM scores (see Table 4) compared with the performances of BERT type embeddings

A comparative evaluation of three generations of Distributional Semantic Models

| Dataset | Static | BERT. F4 | BERT. L4 | BERT. L | Dataset | Static | BERT. F4 | BERT. L4 | BERT. L |
|---|---|---|---|---|---|---|---|---|---|
| *Synonymy* | | | | | *Categorization* | | | | |
| TOEFL | 0.92 | 0.72 | 0.89 | 0.82 | AP | 0.75 | 0.52 | 0.63 | 0.55 |
| ESL | 0.78 | 0.60 | 0.60 | 0.64 | BATTIG | 0.48 | 0.22 | 0.40 | 0.35 |
| *Similarity* | | | | | BATTIG-2010 | 1.00 | 0.67 | 0.77 | 0.73 |
| RG65 | 0.87 | 0.74 | 0.81 | 0.78 | ESSLLI-2008-1a | 0.95 | 0.68 | 0.73 | 0.70 |
| RW | 0.48 | 0.37 | 0.48 | 0.36 | ESSLLI-2008-2b | 0.92 | 0.82 | 0.75 | 0.75 |
| SL-999 | 0.49 | 0.49 | **0.55** | **0.50** | ESSLLI-2008-2c | 0.75 | 0.64 | 0.62 | 0.58 |
| SV-3500 | 0.41 | 0.34 | 0.40 | 0.27 | BLESS | 0.88 | 0.60 | 0.73 | 0.70 |
| WS-353 | 0.71 | 0.61 | 0.62 | 0.57 | *Analogy* | | | | |
| WS-SIM | 0.76 | 0.67 | 0.70 | 0.63 | SAT | 0.34 | 0.24 | 0.24 | 0.21 |
| *Relatedness* | | | | | MSR | 0.68 | **0.76** | 0.69 | 0.68 |
| WS-REL | 0.66 | 0.56 | 0.51 | 0.47 | GOOGLE | 0.76 | 0.38 | 0.66 | 0.64 |
| MTURK | 0.71 | 0.59 | 0.56 | 0.52 | SEMEVAL-2012 | 0.38 | 0.33 | 0.34 | 0.30 |
| MEN | 0.79 | 0.70 | 0.69 | 0.64 | WORDREP | 0.27 | 0.22 | **0.28** | 0.22 |
| TR9856 | 0.17 | 0.13 | 0.14 | 0.13 | BATS | 0.29 | **0.30** | **0.33** | **0.34** |

the BERT.L4 scores respectively of 0.55 and 0.40. The values in Ethayarajh (2019) are even lower than ours. This indicates that the disappointing behavior of our BERT models is not likely to depend on the way the type embeddings were constructed from BERT layers. Research has shown that the performance of contextual DSMs tend to increase with the number of contexts used to build the type vectors. Therefore, we could expect that BERT scores would be higher, if we sampled more sentences, as also reported by Bommasani et al. (2020). However, the advantages brought by more contexts are not always significant: Vulić et al. (2020) argue that the increment produced by sampling 100 sentences instead of 10, as we did, is in most cases marginal.

On the other hand, the performance of the static DSMs used in previous comparisons is often much lower than ours. The best model (i.e., word2vec) in Bommasani et al. (2020) achieves 0.44 in SL-999, 0.68 in WS-353, 0.36 in SV-3500, and 0.68 in RG65. These score are generally worse than the ones obtained by our best static DSMs (see Table 5). Again, the values reported by Ethayarajh (2019) for the static models are even lower. Therefore, we can argue that in those cases BERT "wins" because it is compared with suboptimal static DSMs and the alleged competitiveness or superiority of type embeddings derived from contextual DSMs over static models might be more apparent than real. This resembles the case of the debate between count vs. predict model, in which the advantage of the latter disappears when count DSMs are optimized (Levy et al., 2015). Similarly, when properly tuned, static DSMs are superior to BERT, when tested in out-of-context semantic tasks.

## 6 Representation similarity analysis of semantic spaces

One general shortcoming of the standard way to evaluate DSMs is that it is based on testing their performances on benchmark datasets that, despite their increasing size, only cover a limited portion of a model vocabulary (e.g., the large MEN includes just 751 word types). Apart from few exceptions, the selection criteria of test items are not explicit, and do not take into consideration or do not provide information about important factors such as word frequency and POS. As we saw in the previous section, the variance among datasets is often extremely large, even within the same semantic task, and this might be due to differences in sampling and rating criteria.

We present here an alternative approach that explores the shape of the semantic spaces produced by DSMs with *Representational Similarity Analysis* (RSA) (Kriegeskorte et al., 2008; Kriegeskorte and Kievit, 2013), which has been recently adopted in NLP to analyze the distributed representations of neural networks (Abdou et al., 2019; Abnar et al., 2019; Chrupała and Alishahi, 2019). RSA is a method originally developed in cognitive neuroscience to establish a relationship between a set of stimuli and their brain activations. These are both represented as points in a similarity space (e.g., the stimuli can be represented in terms of their mutual similarity ratings) and are related in terms of the *second-order isomorphism* (Edelman,, 1998) determined by *the similarity of the similarity structure* of the two spaces. Therefore, RSA is a methodology to compare two geometrical representations $R_1$ and $R_2$ of a set of data: The similarity between $R_1$ and $R_2$ depends on how similar the similarity relations among the data according to $R_1$ are to the similarity relations according to $R_2$. As DSMs produce geometrical representations of the lexicon, RSA can be applied to investigate the similarity of their semantic spaces by measuring the correlation between the pairwise similarity relations among the lexical items in different spaces.[11]

We performed RSA on 27 static and contextualized DSMs tested in Sect. 4. This set includes the 300-dimensional models, the PPMI explicit vectors, the RI and RI-perm 2000-dimensional embeddings, which we chose instead of the smaller ones because of their better performance, and the three BERT models. Given a vocabulary $V$ and a DSM $M$, the semantic space representing $V$ corresponds to the matrix **M** whose rows are the vectors of the targets $t \in V$. The RSA of the selected DSMs consisted of the following steps:[12]

---

[11] RSA is akin to the topological analysis used by Brighton and Kirby (2006) and Ren et al. (2020) to measure the convergence of semantic spaces in the emergent communication between artificial agents.

[12] We performed RSA with the Python library Neurora: https://neurora.github.io/NeuroRA/.

– for each DSM $M^k$ generating the semantic space $\mathbf{M}^k$, we built the *representation similarity matrix* $\mathbf{RSM}^k_{|V| \times |V|}$ such that each entry $rsm_{i,j}$ contains the cosine similarity between the vectors $\mathbf{t}_i$ and $\mathbf{t}_j$ of the lexical items $t_i, t_j \in V$;[13]
– for each pair of DSMs $M^1$ and $M^2$, we measured their similarity with the Spearman correlation between $\mathbf{RSM}^1$ and $\mathbf{RSM}^2$.

As the large size of our DSM vocabulary (more than 345, 000 words) made the construction of one global similarity matrix computationally too expensive, we randomly sampled 100 disjoint sets of 1, 000 lexemes and we ran separate analyses on each sample (Sect. 6.1). Further analyses were performed on portions of the DSM vocabulary differing for word frequency (Sect. 6.2) and POS (Sect. 6.3) in the training corpus. The similarity between the semantic spaces produced by two DSMs is the average Spearman correlation between their respective RSMs of the various samples.

### 6.1 RSA of the global semantic spaces

As we can see from the correlation plot in Fig. 5, the global semantic spaces produced by the various DSMs show significant differences, as only few of them have strong correlations. The mean between-DSM Spearman $\rho$ is 0.15 (median = 0.11), with a high variability (sd = 0.21). The similarity between representations is mainly determined by the model type, once again confirming the far greater importance of this factor than the linguistic context in shaping distributional semantic spaces. The predict DSMs (SGNS, CBOW, and FastText) form a dense cluster of mutually similar spaces. They also have a moderate correlation with window-based SVD. Syntax-based SGNS is particularly close to GloVe and LDA. The latter and especially RI are the real outliers of the overall group, since their lexical spaces have very low similarity with any other DSM. Within the same model type, syntax-based spaces predictably tend to be quite similar to narrow window ones. The contextualized BERT models generate semantic spaces that are rather different from static ones. A moderate correlation exists only between BERT.F4 and GloVe, and between BERT.L4, PPMI and some of the SGNS models. Moreover, the BERT.F4 space is quite different from the ones generated by the last layers, probably due to their higher contextualization.

### 6.2 RSA of semantic subspaces selected according to the target frequency

Further RSAs were then performed on subsets of the DSM vocabulary sampled according to their frequency in the training corpus:

*High Frequency (RSA-HF)* the 1,000 most frequent lexemes;

---

[13] Notice that the original RSA method uses dissimilarity matrices whose entries contain dissimilarity measures between the items.

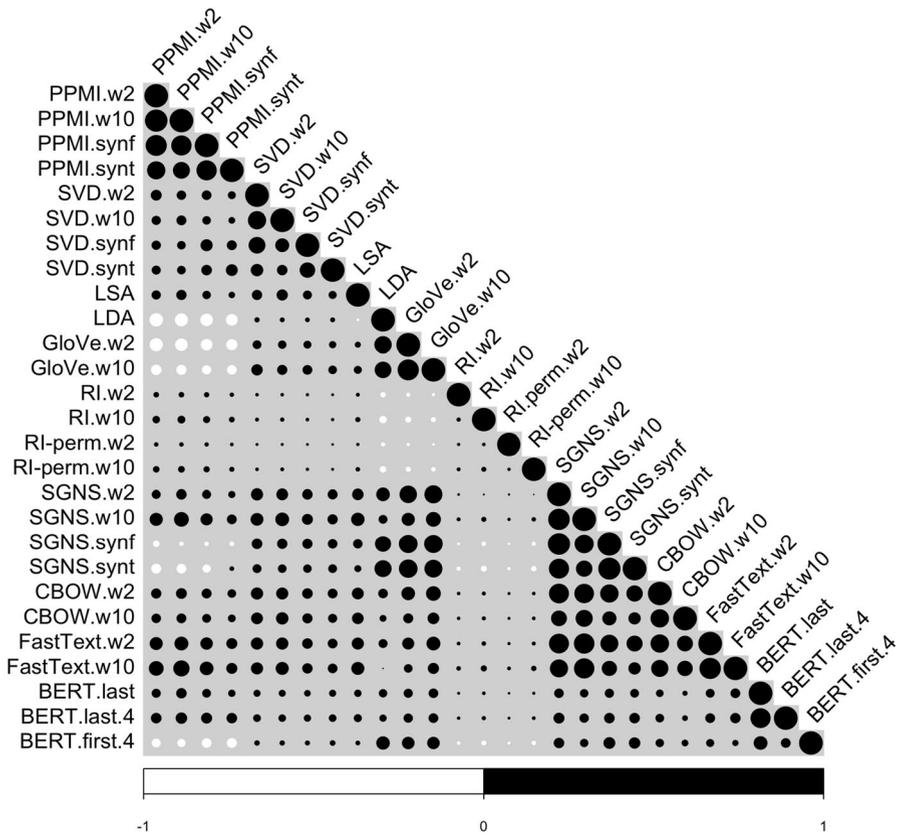

**Fig. 5** Average Spearman correlation between semantic spaces computed with RSA on 100 random samples of 1, 000 words. Dot color marks the correlation sign and dot size its magnitude

*Medium Frequency (RSA-MF)* 10 disjoint random samples of 1,000 lexemes, selected among those with frequency greater than 500, except for the 1,000 most frequent ones;

*Low Frequency (RSA-LF)* 10 disjoint random samples of 1,000 words, selected among those with frequency from 100 up to 500.

The results of these analyses are reported in Fig. 6. It is particularly interesting to notice the great difference in the similarities among semantic spaces depending on the frequency range of the target lexemes. In RSA-HF, most semantic spaces are strongly correlated to each other, apart from few exceptions: The average correlation (mean $\rho$ = 0.44; median = 0.40; sd = 0.22) is in fact significantly higher than the one of the global spaces. Even those models, like RI and LDA, that behave like outliers in the general RSA represent the high frequency semantic space very similarly to the other DSMs. Contextual models also increase their similarity with static ones in the high frequency range. The between-model correlations in

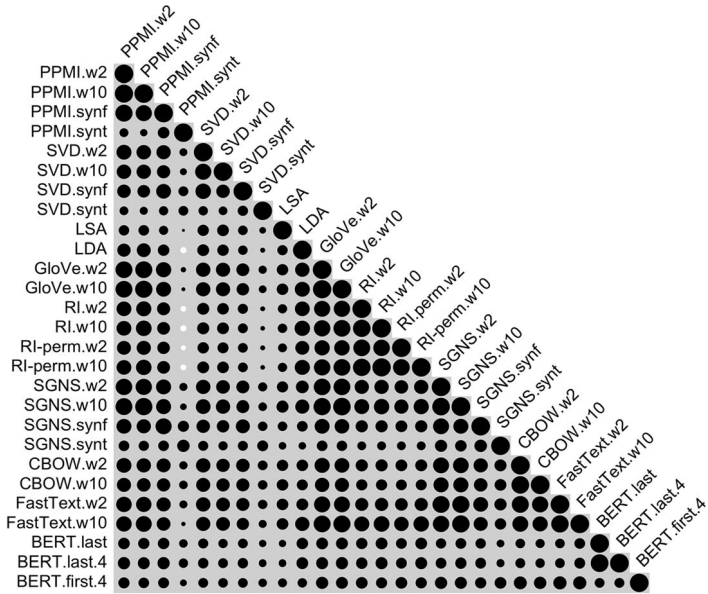

**(a)**

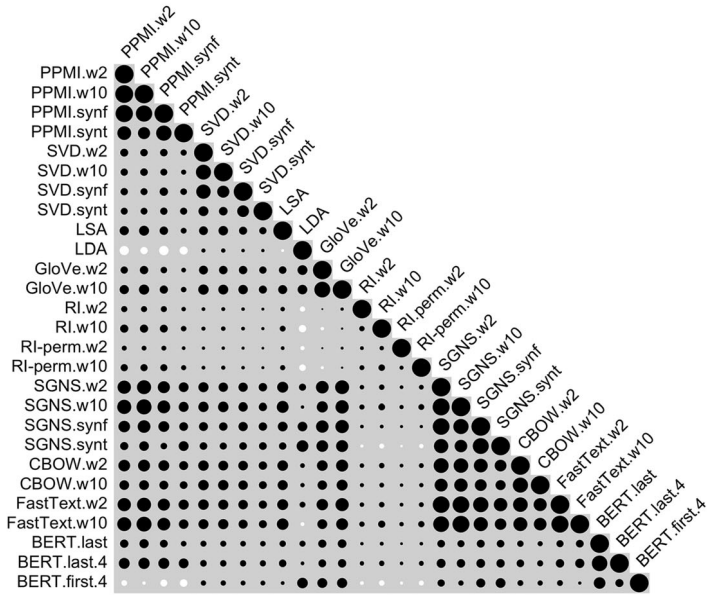

**(b)**

**Fig. 6** Spearman correlation between semantic spaces computed with RSA on **a** high, **b** medium, **c** and low frequency target words. Dot color marks the correlation sign (black positive, white negative), and dot size its magnitude

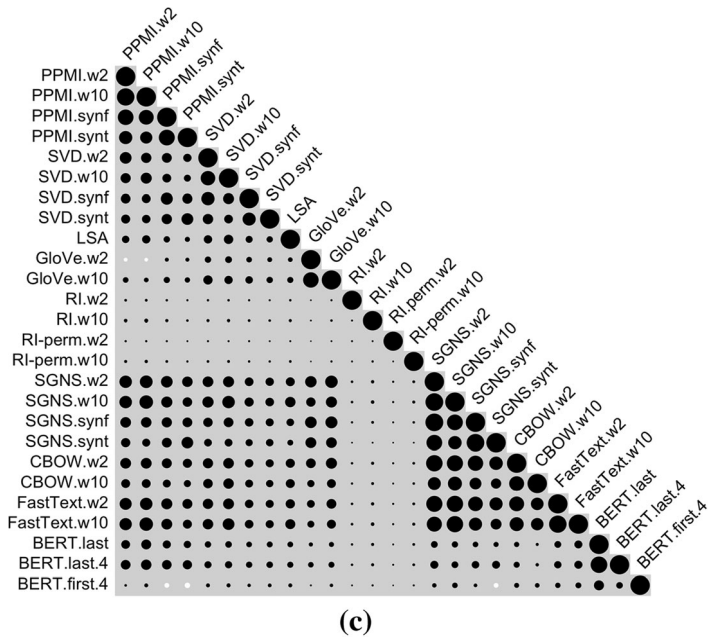

**(c)**

**Fig. 6** continued

RSA-MF (mean $\rho = 0.20$; median = 0.15; sd = 0.20) are significantly lower than RSA-HF (Wilcoxon test: $V = 58843$, $p$-value $< 0.001$). A further decrease occurs with low frequency lexemes (mean $\rho = 0.17$; median = 0.12; sd = 0.19; Wilcoxon test: $V = 10661$, $p$-value $< 0.001$). In this latter case, the effect of model type is particularly strong. The behavior of GloVe is exemplary: Its spaces are very close to the SVD and predict ones for high frequency words, but in the low frequency range they have some moderate correlation only with the latter family of models. Interestingly, in RSA-MF and RSA-LF (window-based) SGNS and FastText are more similar to PPMI than to other count models, probably due to the close link between PPMI and negative sampling proved by Levy and Goldberg (2014c).

It is worth mentioning the peculiar behavior of LDA, which we had to exclude from RSA-LF because most low frequency targets are represented exactly with the same embedding formed by very small values, hence they are not discriminated by the model. We hypothesize that this is due to the way in which word embeddings are defined in Topic Models. As described in Sect. 4.1, given a set of $z_1, \ldots, z_k$ topics, LDA finds the most important words that characterize a topic $z_i$. Each target is then represented with a topic vector $(\phi_1, \ldots, \phi_k)$, such that $\phi_i = p(t|z_i)$. The words that are not relevant to characterize any topic, have low probabilities in all of them. Therefore, these same lexemes are eventually represented by identical topic vectors. The problem is that the size of this phenomenon is actually huge: The LDA.300 model has $305{,}050$ targets with the same topic vector, about $88\%$ of its total vocabulary. Low frequency words are especially affected, probably because they do not appear among the most likely words of any topic, as they occur few times in the documents used as contexts by LDA. This might also explain the systematically low performance of LDA in quantitative evaluation. Moreover, it raises several doubts on its adequacy to build word embeddings in general. Like LSA, Topic Models were originally designed for the semantic analysis of document collections, and were then turned into lexical models on the assumption that, just as documents can be represented with their word distributions, lexical items can be represented with the documents they occur in. However, while this conversion from document to lexical representation works fairly well for LSA, it is eventually problematic for Topic Models.

### 6.3 RSA of semantic subspaces selected according to the target POS

A third group of RSAs was performed on subsets of the DSM vocabulary sampled according to their POS in the training corpus. First, we selected all the words with frequency greater than 500, to avoid the idiosyncrasies produced by low frequency items. Since the DSM targets are not POS-disambiguated, we univocally assigned each selected lexeme to either the noun, verb, or adjective class, if at least $90\%$ of the occurrences of that word in the corpus was tagged with that class. This way, it is likely that the vector of a potentially ambiguous word encodes the distributional properties of its majority POS. At the end of this process we obtained $14{,}893$ common nouns, 7,780 verbs, and 5,311 adjectives. Given the role of target frequency in shaping the semantic spaces, we then split each set in two subsets: High frequency sets are composed by the first 1,000 most frequent targets of each POS, whereas medium frequency sets include the remaining targets. We randomly selected 4 disjoint samples of 1,000 targets from the medium frequency set of each POS (notice that almost all adjectives are represented in these selected samples). The correlation plots of the RSAs on the POS samples are reported in Figs. 9, 10 and 11 in Appendix B.

This analysis shows the effect of frequency in even a clearer way, since for all POS there is a drastic decrease in the model correlations from the high to the medium frequency range, with a symmetric increase in their variability, as shown in Table 6. At the same time, important differences among POS emerge. In the high frequency range, verbs (Wilcoxon test: $V = 15830$, $p$-value $< 0.001$) and adjectives (Wilcoxon test: $V = 19473$, $p$-value $< 0.001$) have a significant higher between-DSM similarity than nouns, and this gap further increases with medium frequency lexical items (verbs instead do not significantly differ from adjectives). This means that the semantic spaces produced by the various DSMs differ much more in the case of nouns than in the case of verbs or adjectives.

**Table 6** Between-DSM correlations with respect to POS and frequency

| POS | Frequency | Mean $\rho$ | Median $\rho$ | sd |
|---|---|---|---|---|
| Nouns | High | 0.53 | 0.51 | 0.17 |
| Verbs | High | 0.56 | 0.53 | 0.17 |
| Adjectives | High | 0.55 | 0.54 | 0.16 |
| Nouns | Medium | 0.28 | 0.25 | 0.23 |
| Verbs | Medium | 0.37 | 0.36 | 0.25 |
| Adjectives | Medium | 0.38 | 0.38 | 0.24 |

## 7 Discussion and conclusions

In this paper, we have presented a comparative evaluation of distributional representations that spans three generations of DSMs, from count to predict models, up to the most recent contextual embeddings. The results of our experiments confirm the highly complex and multifaceted behaviour of DSMs. We summarize here the most important findings of our experiments and we draw some general conclusions.

*The model is the crucial factor* The method to build distributional vectors is the main factor responsible for global and task-based variability of static DSMs.

*Count or predict?* While RI and LDA always lag behind, predict models are prima facie the best performing DSMs. However, in most cases they are not significantly better than SVD or GloVe embeddings, or even explicit PPMI vectors. This means that the alleged superiority of predict models (Baroni et al., 2014; Mandera et al., 2017) is more apparent than real and surely not ubiquitous. When properly optimized, for instance following the suggestions by Levy et al. (2015) like in our models, SVD embeddings are as good as predict ones, at least in most intrinsic tasks. There can be other reasons to prefer predict DSMs, such as incrementality or computational efficiency in processing large corpora, but not the quality itself of the semantic space.

*Static or contextual?* Type vectors derived from the contextual embeddings generated by BERT largely underperform. Recent research has emphasized that their reported good performance makes contextual embeddings highly competitive with static ones (Bommasani et al., 2020; Vulić et al., 2020). Our experiments did not replicate such optimal behavior, and rather prompt the reverse claim: *Static embeddings are still competitive with respect to contextual embeddings (at least BERT ones)*, especially given the simplicity of the former vis-à-vis the complexity of Transformers and the much greater order of magnitude of the computational and data resources needed to train contextual embeddings. In this paper, we have tested just one type of contextual DSMs and a more thorough comparison with the representations produced by other neural language models is surely needed to strengthen and fully generalize this conclusion. Moreover, it is possible that more sophisticated methods of building type embeddings, like the one by Chronis and Erk (2020) with sense clustering, or sampling a larger number of contexts might indeed improve the performances of such models, but this would also further increase the complexity of the type embedding generation process. Therefore, static DSMs still appear as the best and most sustainable option for out-of-context semantic tasks.

*The role of linguistic contexts* The effect of context type is more limited, with the only exception of documents, which are always underperforming. As noticed by Lapesa and Evert (2017), syntax-based contexts are very similar to window-based ones, especially with narrow windows. However, in categorization tasks syntax does provide significant improvements, suggesting that syntactic co-occurrences encode important information about semantic classes. This is consistent with the findings by Chersoni et al. (2017) about the advantage of syntax-based DSMs in semantic tasks modelling predicate-argument typicality. Therefore, syntax-based contexts have an added value over window-based ones, but this is not ubiquitous.

*The role of vector dimensions* Augmenting the embedding size does not produce significant improvements. The only exceptions are RI models, probably because having more dimensions increases the orthogonality of the random vectors.

*Tasks do matter* The effect of tasks on DSM performance is extremely high. This means that, for DSM evaluation to be truly reliable, it should always be performed on a range as large as possible of semantic tasks and datasets.

*Analogies must be handled with care* The "linguistic regularities" (Mikolov et al., 2013c) that can be discovered with the offset method are quite limited, except for morphosyntactic analogies. This confirms the doubts raised in several studies (Linzen, 2016; Rogers et al., 2017; Schluter, 2018; Peterson et al., 2020) about the general soundness of such a method to model analogical similarity. While analogy targets an important aspect of semantic similarity, extreme caution should be exercised in using the analogy completion task and its solution with the offset method as a benchmark to evaluate DSMs.

*Intrinsic vs. extrinsic evaluation* DSM performance on intrinsic tasks correlate with their performance on extrinsic tasks, except for the sequence labelling ones, replicating the findings of Rogers et al. (2018). Differently from what has been sometimes claimed in the literature, this strong correlation indicates that intrinsic evaluation can also be informative about the performance of distributional vectors when embedded in downstream NLP applications. On the other hand, not all extrinsic tasks are equally suitable to evaluate DSMs, as the peculiar behavior of POS, NER and chunking seems to show.

Besides using the traditional approach to DSM evaluation, we have introduced RSA as a task-independent method to compare and explore the representation of the lexical semantic space produced by the various models. In particular, we have found that models, both static and contextualized ones, produce often dramatically different semantic spaces for low frequency words, while for high frequency items the correlation among them is extremely high. This suggests that the main locus of variation among the methods to build distributional representations might reside in how they cope with data sparseness and are able to extract information when the number of occurrences is more limited. In the case of static embeddings, we applied to all DSMs the smoothing and optimization procedure proposed by Levy et al. (2015), but count and predict models still behave very differently in the low frequency range. This indicates that such differences might actually depend on some intrinsic features of the algorithms to build word embeddings, rather than in the setting of their hyperparameters. Overall, these results highlight the strong "instability" (Antoniak and Mimno, 2018) of distributional semantic spaces, especially when the target frequency is not high: Models can produce substantially divergent representations of the lexicon, even when trained on the same corpus data with highly comparable settings. The DSM instability at low frequencies is particularly critical for their application in low-resource settings in linguistics, digital humanities (e.g., diachronic research), and for studies on language acquisition. Investigating which DSMs are most suited to represent the content of low-frequency lexemes is an important and still open research question.

Significant variations also occur at the level of POS too. Quite unexpectedly, the category where DSM spaces differ most are nouns. This finding deserves future investigations to understand the source of such differences, and to carry out more fine-grained analyses within nouns (e.g., zooming in on particular subclasses, such as abstract vs. concrete ones). These analyses reveal that frequency and POS strongly affect the shape of distributional semantic spaces and must therefore be carefully considered when comparing DSMs.

We conclude this paper with one last observation. In more than thirty years, distributional semantics has undoubtedly been making enormous progresses, since the performance of DSMs as well as the range of their applications have greatly increased. On the other hand, we might argue that this improvement is mostly due to better optimized models or to a more efficient processing of huge amounts of training data, rather than to a real breakthrough in the methods to distil semantic information from distributional data. In fact, under closer scrutiny, the most recent and sophisticated algorithms have not produced dramatic advances with respect to more traditional ones. This raises further general questions that we leave to future research: Are we reaching the limits of distributional semantics? How to fill the gap between current computational models and the human ability to learn word meaning from the statistical analysis of the linguistic input?

**Authors contributions** AL: experiment design and coordination, data analysis, paper drafting; MS: experiment design and coordination, support in writing the paper; PJ: processing the corpora, training the DSMs, intrinsic evaluation, support in writing the paper; ACG: extrinsic evaluation, support in writing the paper; MM: Representational Similarity Analysis, support in writing the paper.

# Appendixes

## A: Statistical analyses of static DSMs

See Figs. 7 and 8, Table 7, 8, 9 and 10.

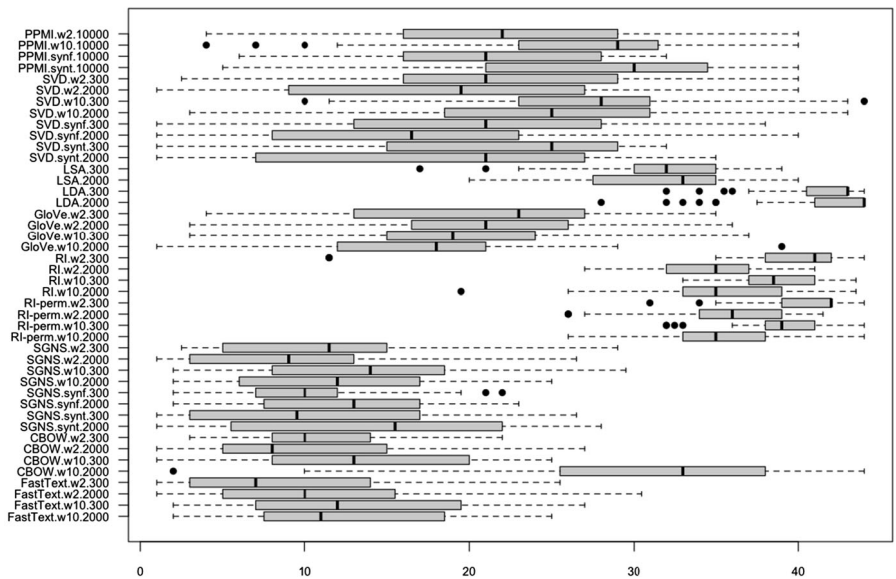

**Fig. 7** Rank distribution of the performance of the 44 static DSMs on the 25 intrinsic and 8 extrinsic datasets (cf. Sect. 4.2)

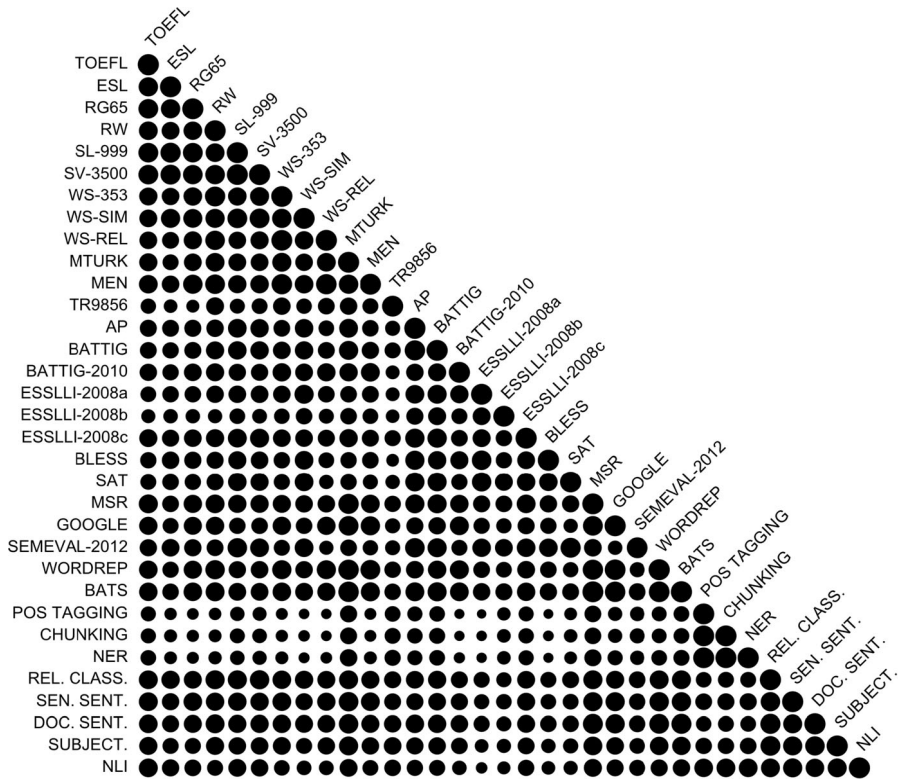

**Fig. 8** Spearman correlation between datasets. The bigger is the dot size the higher is the correlation

**Table 7** P-values of Dunn's tests for multiple comparisons of model types

|         | CBOW   | FastText | GloVe  | LDA    | LSA    | PPMI    | RI     | RI-perm | SGNS   |
|---------|--------|----------|--------|--------|--------|---------|--------|---------|--------|
| FastText | 0.03 | | | | | | | | |
| GloVe   | **1**  | <0.001   | | | | | | | |
| LDA     | <0.001 | <0.001   | <0.001 | | | | | | |
| LSA     | <0.001 | <0.001   | <0.001 | <0.001 | | | | | |
| PPMI    | <0.001 | <0.001   | 0.036  | <0.001 | 0.008  | | | | |
| RI      | <0.001 | <0.001   | <0.001 | **0.508** | **0.379** | <0.001 | | | |
| RI-perm | <0.001 | <0.001   | <0.001 | **1**  | **0.068** | <0.001 | **1** | | |
| SGNS    | 0.009  | **1**    | <0.001 | <0.001 | <0.001 | <0.001  | <0.001 | <0.001 | |
| SVD     | 0.037  | <0.001   | **1**  | <0.001 | <0.001 | **1**   | <0.001 | <0.001 | <0.001 |

**Table 8** P-values of Dunn's tests for multiple comparisons of context types

|  | syntax.filtered | syntax.typed | document | window.10 |
|---|---|---|---|---|
| syntax.typed | **1** |  |  |  |
| document | <0.001 | <0.001 |  |  |
| window.10 | <0.001 | <0.001 | <0.001 |  |
| window.2 | <0.001 | **0.356** | <0.001 | 0.001 |

**Table 9** Dunn's tests for multiple comparisons of model types for each semantic task

|  | Task | PPMI | SVD | LSA | LDA | GloVe | RI |
|---|---|---|---|---|---|---|---|
| *SVD* | Synonymy | ○ |  |  |  |  |  |
|  | Similarity | ● |  |  |  |  |  |
|  | Relatedness | ○ |  |  |  |  |  |
|  | Categorization | ○ |  |  |  |  |  |
|  | Analogy | ○ |  |  |  |  |  |
|  | Extrinsic | ● | *SVD* |  |  |  |  |
| *LSA* | Synonymy | ○ | ○ |  |  |  |  |
|  | Similarity | ○ | ● |  |  |  |  |
|  | Relatedness | ○ | ○ |  |  |  |  |
|  | Categorization | ○ | ● |  |  |  |  |
|  | Analogy | ○ | ○ |  |  |  |  |
|  | Extrinsic | ○ | ○ | *LSA* |  |  |  |
| *LDA* | Synonymy | ○ | ● | ○ |  |  |  |
|  | Similarity | ○ | ● | ○ |  |  |  |
|  | Relatedness | ○ | ● | ○ |  |  |  |
|  | Categorization | ● | ● | ○ |  |  |  |
|  | Analogy | ● | ● | ○ |  |  |  |
|  | Extrinsic | ● | ○ | ○ | *LDA* |  |  |
| *GloVe* | Synonymy | ○ | ○ | ○ | ● |  |  |
|  | Similarity | ○ | ○ | ○ | ● |  |  |
|  | Relatedness | ○ | ○ | ○ | ○ |  |  |
|  | Categorization | ○ | ○ | ○ | ● |  |  |
|  | Analogy | ○ | ○ | ● | ● |  |  |
|  | Extrinsic | ○ | ● | ● | ● | *GloVe* |  |
| *RI* | Synonymy | ○ | ● | ○ | ○ | ● |  |
|  | Similarity | ○ | ● | ○ | ○ | ● |  |
|  | Relatedness | ○ | ● | ○ | ○ | ○ |  |
|  | Categorization | ● | ● | ○ | ○ | ● |  |
|  | Analogy | ● | ● | ○ | ○ | ● |  |
|  | Extrinsic | ● | ○ | ○ | ○ | ● |  |

**Table 9** continued

| | Task | PPMI | SVD | LSA | LDA | GloVe | RI | | | |
|---|---|---|---|---|---|---|---|---|---|---|
| *RI-perm* | Synonymy | ○ | ● | ○ | ○ | ● | ○ | | | |
| | Similarity | ○ | ● | ○ | ○ | ● | ○ | | | |
| | Relatedness | ○ | ● | ○ | ○ | ○ | ○ | | | |
| | Categorization | ● | ● | ○ | ○ | ● | ○ | | | |
| | Analogy | ● | ● | ○ | ○ | ● | ○ | | | |
| | Extrinsic | ● | ○ | ○ | ○ | ● | ○ | *RI-perm* | | |
| *SGNS* | Synonymy | ● | ○ | ○ | ● | ○ | ● | ● | | |
| | Similarity | ● | ○ | ● | ● | ○ | ● | ● | | |
| | Relatedness | ○ | ○ | ○ | ● | ○ | ● | ● | | |
| | Categorization | ● | ○ | ● | ● | ○ | ● | ● | | |
| | Analogy | ○ | ● | ● | ● | ○ | ● | ● | | |
| | extrinsic | ● | ● | ● | ● | ○ | ● | ● | *SGNS* | |
| *CBOW* | synonymy | ○ | ○ | ○ | ○ | ○ | ○ | ○ | ○ | |
| | Similarity | ● | ○ | ● | ● | ○ | ● | ● | ○ | |
| | Relatedness | ● | ○ | ○ | ● | ○ | ● | ● | ○ | |
| | Categorization | ○ | ○ | ○ | ● | ○ | ● | ● | ○ | |
| | Analogy | ○ | ○ | ○ | ● | ○ | ● | ● | ○ | |
| | Extrinsic | ○ | ● | ● | ● | ○ | ● | ● | ○ | *CBOW* |
| *FastText* | Synonymy | ○ | ○ | ○ | ● | ○ | ● | ● | ○ | ○ |
| | Similarity | ● | ○ | ● | ● | ○ | ● | ● | ○ | ○ |
| | Relatedness | ● | ● | ● | ● | ● | ● | ● | ○ | ○ |
| | Categorization | ○ | ○ | ● | ● | ○ | ● | ● | ○ | ○ |
| | Analogy | ● | ● | ● | ● | ○ | ● | ● | ○ | ○ |
| | Extrinsic | ● | ● | ● | ● | ○ | ● | ● | ○ | ○ |

Black dots mark significantly different models ($p < 0.05$)

**Table 10** Dunn's tests for multiple comparisons of context types for each semantic task

|  | Task | window.2 |  |  |  |
| --- | --- | --- | --- | --- | --- |
| window.10 | Synonymy | ○ |  |  |  |
|  | Similarity | ○ |  |  |  |
|  | Relatedness | ○ |  |  |  |
|  | Categorization | ○ |  |  |  |
|  | Analogy | ○ |  |  |  |
|  | Extrinsic | ○ | window.10 |  |  |
| syntax.filtered | Synonymy | ○ | ○ |  |  |
|  | Similarity | ○ | ● |  |  |
|  | Relatedness | ○ | ○ |  |  |
|  | Categorization | ● | ○ |  |  |
|  | Analogy | ○ | ● |  |  |
|  | Extrinsic | ○ | ○ | syntax.filtered |  |
| syntax.typed | Synonymy | ○ | ○ | ○ |  |
|  | Similarity | ○ | ○ | ○ |  |
|  | Relatedness | ○ | ○ | ○ |  |
|  | Categorization | ● | ● | ○ |  |
|  | Analogy | ○ | ○ | ○ |  |
|  | Extrinsic | ○ | ○ | ○ | syntax.typed |
| document | Synonymy | ● | ○ | ● | ● |
|  | Similarity | ● | ● | ● | ● |
|  | Relatedness | ● | ● | ● | ● |
|  | Categorization | ● | ● | ● | ● |
|  | Analogy | ● | ● | ● | ● |
|  | Extrinsic | ● | ● | ● | ● |

Black dots mark significantly different contexts ($p < 0.05$)

## B: RSAs on target samples selected according to their POS

See Figs. 9, 10 and 11.

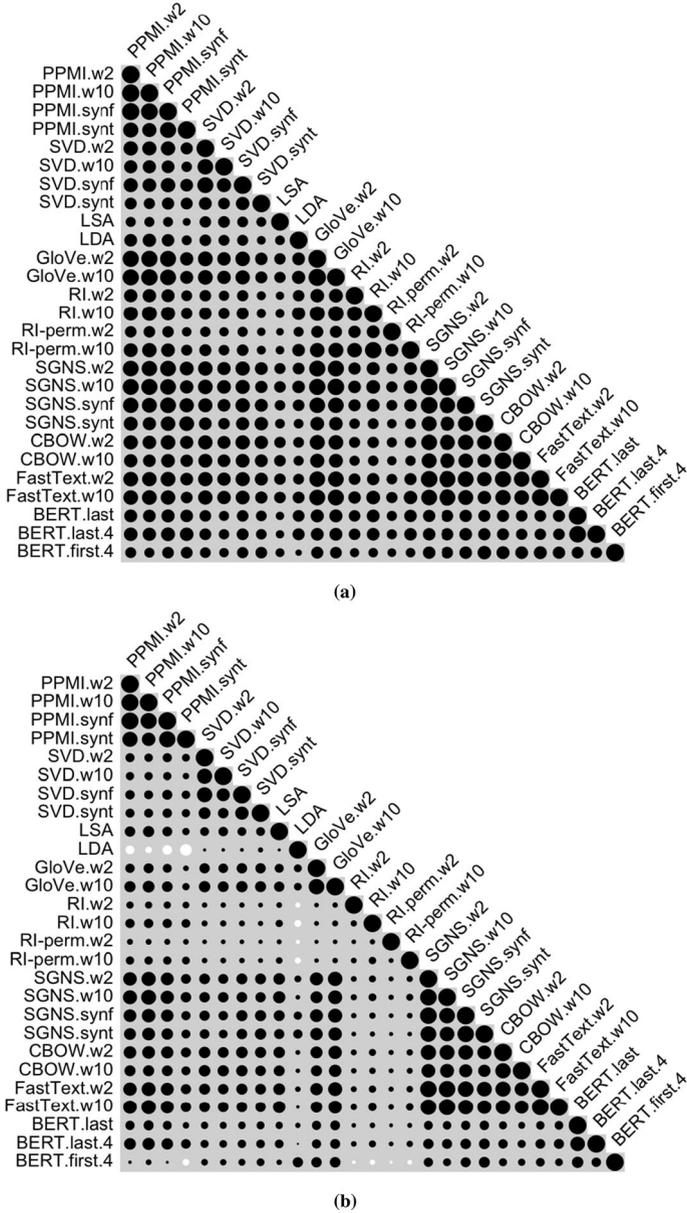

(a)

(b)

**Fig. 9** Spearman correlation between semantic spaces computed with RSA on high frequency (**a**) and medium frequency (**b**) nouns. Dot color marks the correlation sign (black positive, white negative), and dot size its magnitude

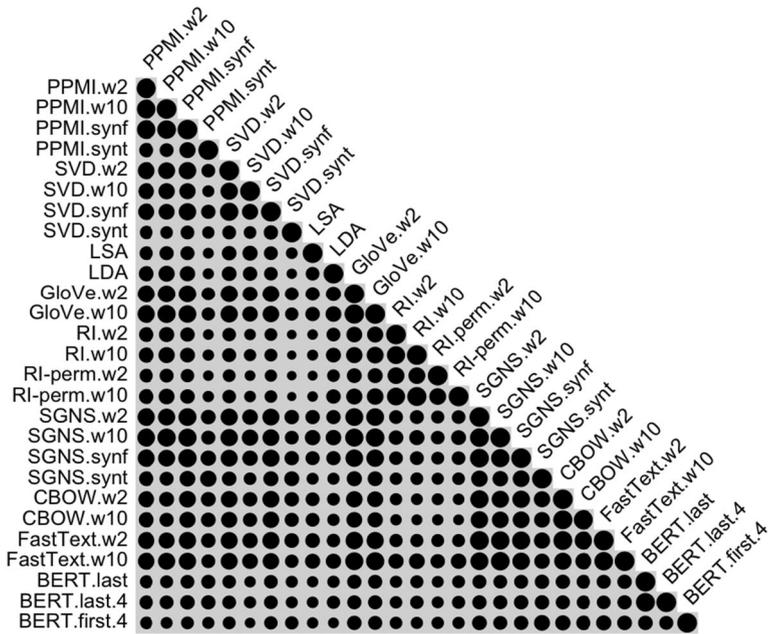

(a)

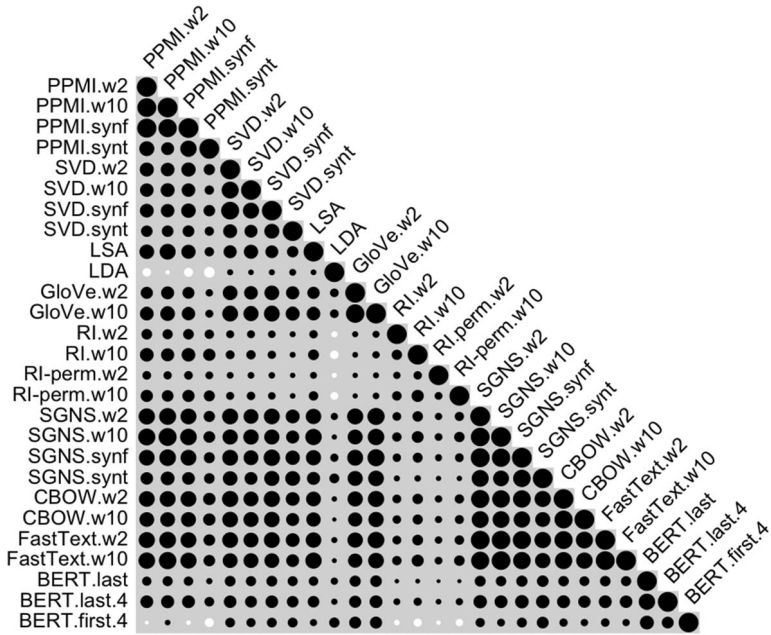

(b)

**Fig. 10** Spearman correlation between semantic spaces computed with RSA on high frequency (**a**) and medium frequency (**b**) verbs. Dot color marks the correlation sign (black positive, white negative), and dot size its magnitude

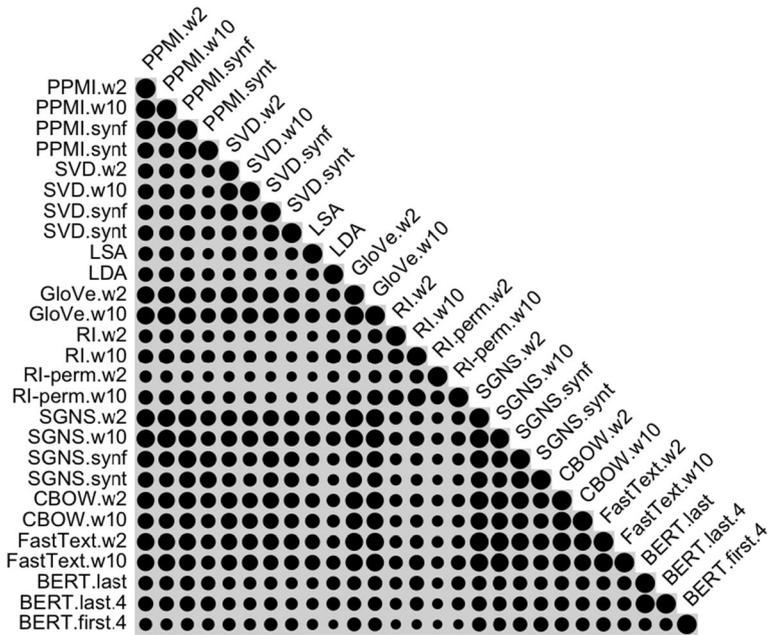

**(a)**

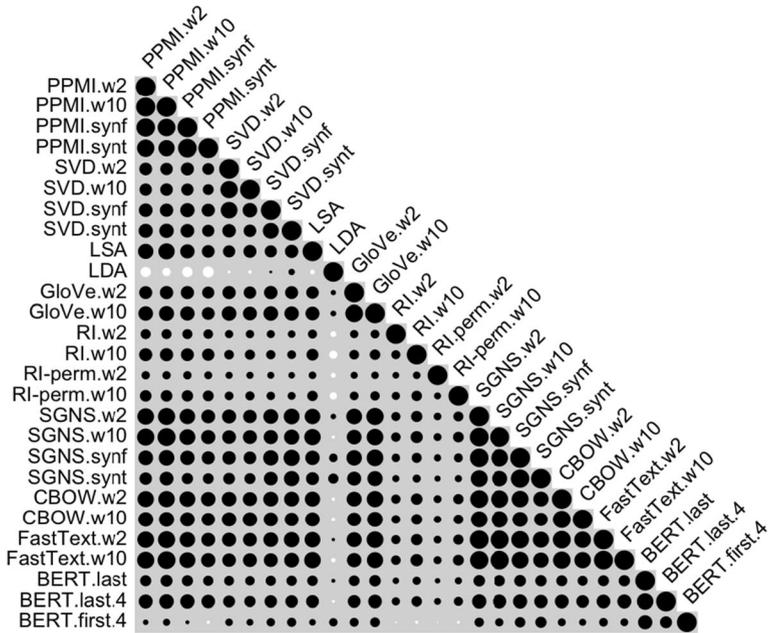

**(b)**

**Fig. 11** Spearman correlation between semantic spaces computed with RSA on high frequency (**a**) and medium frequency (**b**) adjectives. Dot color marks the correlation sign (black positive, white negative), and dot size its magnitude